\documentclass[letterpaper]{article} 
\usepackage{aaai23}  
\usepackage{times}  
\usepackage{helvet}  
\usepackage{courier}  
\usepackage[hyphens]{url}  
\usepackage{graphicx} 
\usepackage{natbib}  
\usepackage{caption} 
\usepackage{algorithm}
\usepackage{algorithmic}

\usepackage{newfloat}
\usepackage{listings}
\DeclareCaptionStyle{ruled}{labelfont=normalfont,labelsep=colon,strut=off} 
\floatstyle{ruled}
\newfloat{listing}{tb}{lst}{}
\floatname{listing}{Listing}
\usepackage{amsfonts}
\usepackage{amsmath}
\usepackage{subcaption}

\title{The Unreasonable Effectiveness of Deep Evidential Regression}
\author{
    Nis Meinert, \textsuperscript{\rm 1}\\
    Jakob Gawlikowski, \textsuperscript{\rm 2}\\
    Alexander Lavin \textsuperscript{\rm 1}
}
\affiliations{

    \textsuperscript{\rm 1}Pasteur Labs, 19 Morris Avenue, Brooklyn Navy Yard Building 128, Brooklyn, NY 11205, USA\\
    \textsuperscript{\rm 2}German Aerospace Center, Institute of Data Science, M\"alzerstra\ss{}e 3-5, 07745 Jena, Germany\\
    nis.meinert@simulation.science, jakob.gawlikowski@dlr.de, lavin@simulation.science
}

\usepackage{bibentry}

\begin{document}

\maketitle

\begin{abstract}
There is a significant need for principled uncertainty reasoning in machine learning systems as they are increasingly deployed in safety-critical domains.
A new approach with uncertainty-aware regression-based neural networks (NNs), based on learning evidential distributions for aleatoric and epistemic uncertainties, shows promise over traditional deterministic methods and typical Bayesian NNs, notably with the capabilities to disentangle aleatoric and epistemic uncertainties.
Despite some empirical success of Deep Evidential Regression (DER), there are important gaps in the mathematical foundation that raise the question of why the proposed technique seemingly works.
We detail the theoretical shortcomings and analyze the performance on synthetic and real-world data sets, showing that Deep Evidential Regression is a heuristic rather than an exact uncertainty quantification.
We go on to discuss corrections and redefinitions of how aleatoric and epistemic uncertainties should be extracted from NNs.
\end{abstract}

\section{Introduction}
Using neural networks (NNs) for regression tasks is one of the main applications of modern machine learning.
Given a dataset of $(\vec{x}_i, \vec{y}_i)$ pairs, the typical objective is to train a NN $\vec{f}_i \equiv \vec{f}(\vec{x}_i | \boldsymbol{\omega})$ w.r.t.\ $\boldsymbol{\omega}$ such that a given loss $\mathcal{L}(\vec{y}_i, \vec{f}_i) \equiv \mathcal{L}_i(\boldsymbol{\omega})$ becomes minimal for each $(\vec{x}_i, \vec{y}_i)$ pair.
Traditional regression-based NNs are designed to output the regression target, a.k.a., the prediction for $\vec{y}_i$, directly which allows a subsequent minimization, for example of the sum of squares:
\begin{equation}
    \min\limits_{\boldsymbol{\omega}} \sum\limits_i \mathcal{L}_i(\boldsymbol{\omega}) = \min\limits_{\boldsymbol{\omega}} \sum\limits_i \left( \vec{y}_i - \vec{f}(\vec{x}_i | \boldsymbol{\omega}) \right)^2.
\end{equation}
Technically, this is nothing but a fit of a model $\vec f$, parameterized with $\boldsymbol{\omega}$, w.r.t.\ $\sum_i \mathcal{L}_i$ to data.
As with any fit, the model has to find a balance between being too specific (over-fitting) and being too general (under-fitting).
In machine learning this balance is typically evaluated by analyzing the trained model on a separated part of the given data which was not seen during training.
In practice, no model will be able to describe this evaluation sample perfectly and deviations can be categorized into two groups: \textit{aleatoric} and \textit{epistemic} uncertainties \citep{hora96,kiureghian09,kendall17,huellermeier21}.
The former quantifies system stochasticity such as observation and process noise, and the latter is model-based or subjective uncertainty due to limited data.

The field has largely focused on Bayesian NN approaches that use Monte Carlo sampling and other approximate inference techniques to estimate uncertainty of deep NN models.
A different approach to uncertainty-aware NNs may be useful to more efficiency quantify, and also to disentangle, the several types of uncertainties:
\textit{Deep Evidential Regression} (DER) aims to simultaneously predict both uncertainty types in a single forward pass without sampling or utilization of out-of-distribution data, based on learning evidential distributions for aleatoric and epistemic uncertainties \citep{amini20}.
Yet only with simple empirical demonstrations on univariate regression tasks, this technique has already been applied and recommended in medical and other safety critical applications \citep{zhijian21,soleimany21,cai21,chen21,singh22,petek22,li22}.
With an alternative derivation and experimentation, we identify theoretical shortcomings that do not justify the empirical results let alone the assumed reliability in practice --- it can be vital to understand to what degree the uncertainty estimations are trustworthy.

In the following, we resolve the supposedly unreasonable effectiveness and relate it to convergence patterns of the predicted uncertainties.
Furthermore, we propose re-definitions of the original approach and discuss generalizations.

\subsection{Disentangling Aleatoric and Epistemic Uncertainties}
The residual between the genuine and the predicted value of an imperfect model can be disentangled into an \textit{aleatoric} and \textit{epistemic} contribution.
In theory, the former is related to the noise level in the data and, typically, does not depend on the sample size.
In contrast, the latter scales with the sample size, and either allows the model to be pulled towards the observed distribution if the sample size is increased in this region, or allows the fit of a more complex model and thus decreasing under-fitting in general.
In practice, however, both types can be non-trivially correlated and disentangling both without relying on a-priori assumptions is often ambiguous.
Even if both types are uncorrelated, observing a single deviation from the ground truth $\delta(x)$ at point $x$, linear error propagation, $\delta^2(x) = u_\text{al}^2(x) + u_\text{ep}^2(x)$, shows that a separation of the aleatoric $u_\text{al}(x)$ and the epistemic uncertainty $u_\text{ep}(x)$ is impossible to do point-wise without assuming a certain level of smoothness in $x$ and prior knowledge about at least one of the uncertainties.

For now take the simplest setting: univariate, point-wise normal distributed data, i.e., $(x_i, y_i + \epsilon_i)$ where the noise, $\epsilon_i \sim \mathcal{N}(0, \sigma^2_i)$, is heteroscedastic, i.e., not necessarily equally distributed for all $x_i$ and our goal is to predict  the mean and the noise level depending on the input $x_i$.
In a Bayesian framework, this corresponds to taking a normal-inverse-gamma distribution, $\operatorname{NIG}(\mu, \sigma^2 | \boldsymbol{m})$ with $\boldsymbol{m}=(\gamma, \nu, \alpha, \beta)$, as the conjugated prior of a normal distribution with unknown mean $\mu$ and variance $\sigma^2$.
Integrating out the nuisance parameters, Bayesian inference yields that the likelihood of an observation $y_i$ given $\boldsymbol{m}$ follows a $t$-distribution with $2\alpha_i$ degrees of freedom \citep{gelman13,amini20},
\begin{equation}
    \label{eq:studentt}
    L_i^\text{NIG} = \operatorname{St}_{2\alpha_i}\!\left( y_i \middle| \gamma_i, \frac{\beta_i (1 + \nu_i)}{\nu_i \alpha_i} \right).
\end{equation}
If $\boldsymbol{m}$ is known, it is reasonable to define the prediction of $y_i$ as $\mathbb{E}[\mu_i] = \gamma_i$, and the aleatoric and epistemic uncertainties $u_\text{al}$ and $u_\text{ep}$, respectively, as:
\begin{align}
    \label{eq:uncertainties}
    u^2_\text{al} &\equiv \mathbb{E}[\sigma_i^2] = \beta_i / (\alpha_i - 1) &
    u^2_\text{ep} &\equiv \operatorname{var}[\mu_i] = \mathbb{E}[\sigma_i^2] / \nu_i \,.
\end{align}

\subsection{State of the Art and its Issues}
\label{sec:sotaissue}
One may learn Eq.~\eqref{eq:studentt} from data by minimizing
\begin{equation}
    \label{eq:loss}
    \mathcal{L}_i(\boldsymbol{\omega}) = -\log L_i^\text{NIG}(\boldsymbol{\omega}) + \lambda \underbrace{|y_i - \gamma_i|\Phi}_{\mathcal{L}_i^\text{R}(\boldsymbol{\omega})},
\end{equation}
where $\boldsymbol{m} = \operatorname{NN}(\boldsymbol{\omega})$ is given by a NN, $\lambda$ is a tunable hyperparameter, and $\Phi=2\nu_i + \alpha_i$ is the \textit{total evidence}.
The definition of the total evidence is motivated by the fact that taking a $\text{NIG}$ distribution as a conjugated prior corresponds to assuming prior knowledge about the mean and the variance extracted from $\nu_i$ virtual measurements of the former and $2\alpha_i$ virtual measurements for the latter \citep{gelman13}.
For consistency we have adopted the definition of $\Phi$ as shown above but note that using $\Phi=\nu_i + 2\alpha_i$ would actually be better motivated as pointed out first by \citet{meinert21} and already used as such in \citet{zhijian21}.

The issue with the existing approach, as proposed by \citet{amini20} and shown in Eq.~\eqref{eq:loss}, is that minimizing $\mathcal{L}_i(\boldsymbol{\omega})$ w.r.t.\ $\boldsymbol{\omega}$ is insufficient to find $\boldsymbol{m}$.
This is obvious by noting the overparameterization in Eq.~\eqref{eq:studentt} that is not resolved by adding the regularization $\mathcal{L}_i^\text{R}(\boldsymbol{\omega})$, and can also be shown with some mathematical rigor by finding
\begin{equation}
    \frac{\partial}{\partial \nu_i} \log L_i^\text{NIG} = 0 \quad \text{if } \beta_i(\nu_i) \propto \frac{1}{1 + \nu_i^{-1}},
\end{equation}
to reveal that $\log L_i^\text{NIG}$ does not depend on $\nu_i$ and thus $\mathcal{L}_i(\boldsymbol{\omega})$ is minimized, independent of the data, by sending $\nu_i \to 0$ and, for instance, following a path along $\beta_i(\nu_i) = 1 / (1 + \nu_i^{-1})$.
Clearly, a loss function which can be minimized independent of data is non-informative for $\boldsymbol{m}$ and thus cannot be used for evaluating Eqs.~\eqref{eq:uncertainties}.
The underlying reason for this overparameterization, which allows for choosing a path which minimizes the loss independent of data, is the fact that $L_i^\text{NIG}$ is by definition a projection of the NIG distribution and thus unable to unfold all of its degrees of freedom unambiguously.
This ambiguity stays unresolved due to a missing additional constraint for $\beta_i$ in the regularizer $\mathcal{L}_i^\text{R}$.

Curiously, if applied to synthetic and real-world data, the aforementioned approach does yield reasonable results \citep{amini20,zhijian21,cai21,soleimany21,singh22,petek22,li22}.
In particular, areas with a low data density during training that result in a large model uncertainty during inference are identified with large values in the epistemic uncertainty if estimated according to Eq.~\eqref{eq:uncertainties} and low values elsewhere.

\section{The Unreasonable Effectiveness}
In this section we show why Deep Evidential Regression (DER) can yield reasonable results in practice despite its issues --- i.e., why it is unreasonably effective --- and we identify and redefine the key components of extracting disentangled uncertainties.

\subsection{Measuring Aleatoric Uncertainty by Convergence Speed}
\begin{figure*}
    \centering
    \begin{subfigure}[b]{.32\textwidth}
        \centering
        \includegraphics[width=\textwidth]{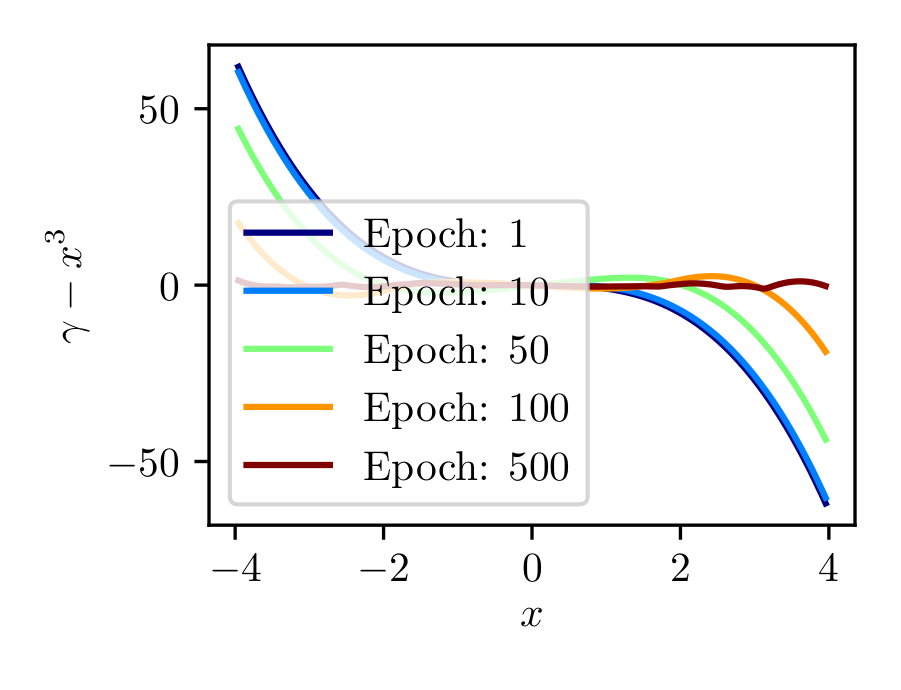}
        \caption{Evolution of residual.}
        \label{fig:error_evol}
    \end{subfigure}
    \hfill
    \begin{subfigure}[b]{.32\textwidth}
        \centering
        \includegraphics[width=\textwidth]{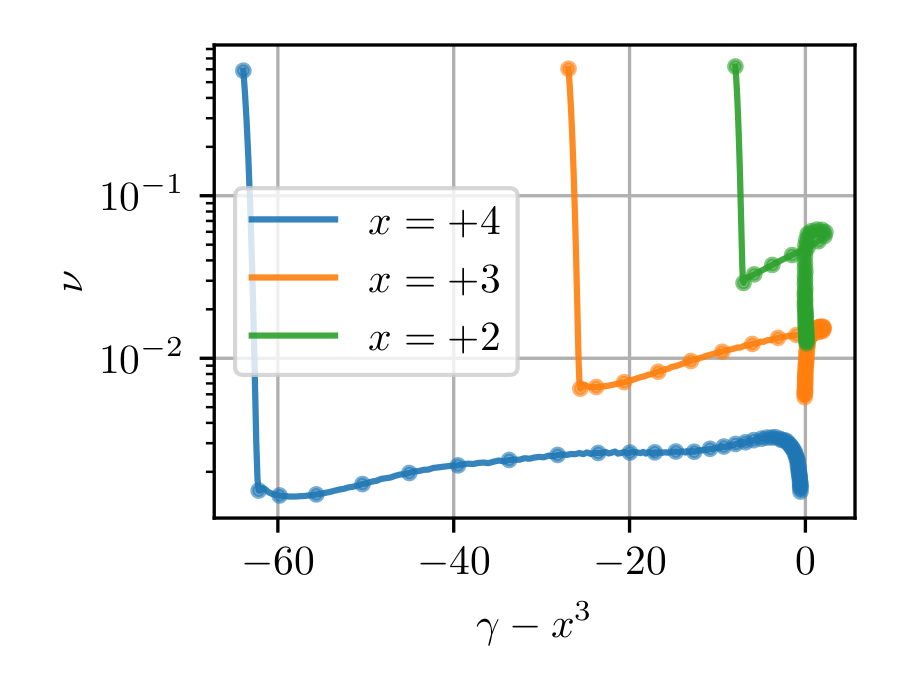}
        \caption{Evolution of $\nu$ w.r.t.\ residual.}
        \label{fig:nu_evol}
    \end{subfigure}
    \hfill
    \begin{subfigure}[b]{.32\textwidth}
        \centering
        \includegraphics[width=\linewidth]{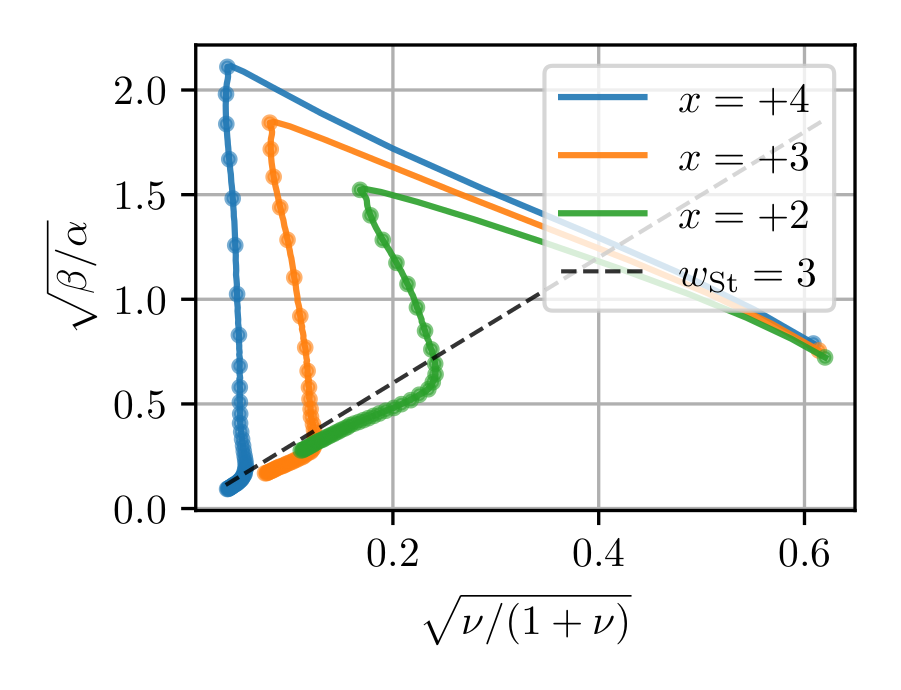}
        \caption{Evolution of factors of $w_\text{St}$.}
        \label{fig:ba_nu_evol}
    \end{subfigure}
    \caption{Evolution of parameters during training. In total, 50 independently trained samples are averaged in each epoch. Dots are placed every 10 epochs to indicate the speed and direction of the convergence. (Left) Evolution of the residual $\gamma_i - x^3$ for all $x \in [-4, 4]$. (Center) Evolution of $\nu_i$ w.r.t.\ residual at three different $x$-positions. (Right) Evolution of $\sqrt{\beta_i / \alpha_i}$ and $\sqrt{\nu_i / (1 + \nu_i}$ at different $x$ positions. The ratio of the quantities is the estimation of the width of a $t$-distribution, $w_\text{St}$. A constant width $w_\text{St}=3$ is indicate by a dashed line and referred to in the text as the \textit{valley}.}
\end{figure*}
In order to analyze the convergence behavior of DER we set up the same synthetic experiment that was used by \citet{amini20}.
A shallow, fully connected NN with a single input and four output neurons, $(\theta_{i,1}, \theta_{i,2}, \theta_{i,3}, \theta_{i,4})$, is used to predict for a given $x_i$ the parameters of a $\text{NIG}$ distribution $\boldsymbol{m} = (\gamma_i, \nu_i, \alpha_i, \beta_i)$,
\begin{equation}
\begin{aligned}
    \gamma_i &= \theta_{i,1}, &
    \nu_i &= \operatorname{softplus}(\theta_{i,2}),\\
    \alpha_i &= \operatorname{softplus}(\theta_{i,3}) + 1, &
    \beta_i &= \operatorname{softplus}(\theta_{i,4}),
\end{aligned}
\end{equation}
where $\operatorname{softplus}(\bullet) = \log(1 + \exp(\bullet))$ enforces non-negative values.

A data sample is generated by generating 1k pairs $(x_i, x_i^3 + \epsilon_i)$ with $\epsilon_i \sim \mathcal{N}(0, \sigma^2=9)$ and $x_i$ uniformly distributed on $[-4, +4]$.
The NN is trained with $\lambda=0.01$ and the Adam optimizer with an initial learning rate of $5 \times 10^{-4}$ for 500 epochs.
We repeat this experiment $50$ times with different seeds for the initialization of the NN.
The results of the first nine independently trained NNs are shown in the Appendix in Fig.~\ref{fig:sota} and reproduce the findings of \citet{amini20} despite a certain variance among the samples:
On the larger interval $x \in [-7, +7]$, the model predicts a large epistemic uncertainty in regions where it has never seen data during training.

In Fig.~\ref{fig:error_evol} we show the convergence of $\gamma_i(x)$ in time and find that the convergence speed differs across the interval in $x$.
Large residuals induce large gradients for $\nu_i$ in the regularizer,
\begin{equation}
    \frac{\partial \mathcal{L}_i^\text{R}}{\partial \nu_i} \propto \lambda |y_i - \gamma_i|,
\end{equation}
and thus push $\nu_i$ towards smaller values faster, as shown in Fig.~\ref{fig:nu_evol}.
Our studies show that residuals of the very first epochs predominantly define how small $\nu_i$ will eventually become since the magnitude of its gradient gradual decreases the better the model gets.

In the Appendix in Fig.~\ref{fig:alpha_evol} and Fig.~\ref{fig:alpha} we show the corresponding analysis for $\alpha_i$ and find a similar behavior, however, here a second gradient from $\log L_i^\text{NIG}$ is conflicting with the gradient of the regularizer which slows down convergence and eventually stop at values $\alpha_i \approx 2$ for $-4 < x_i < +4$.

In summary: our first key insight to understand DER is that the point-wise convergence speed of the model is used as a proxy for the epistemic uncertainty, which arguably stretches the canonical definition.
As a consequence, the numerical values of the epistemic uncertainty cannot be interpreted as canonical Bayesian or Frequentist uncertainty estimations and only relative changes are conclusive.
In that sense, DER is a heuristic that has proven to yield decent results in practice.

\subsection{How to not extend DER}
It is tempting to generalize this approach and use it for arbitrary functions $f(\vec{x}|\boldsymbol{\omega})$, that sufficiently describes the data distribution, by adding a regularizer to the NLL part of the loss function,
\begin{equation}
    \mathcal{L}_i(\boldsymbol{\omega}, \nu_i) = -\log f(\boldsymbol{\omega}) + \lambda |\vec{y}_i - \vec{\gamma}_i| \nu_i.
\end{equation}
However, empirically we find that optimizers, especially those that include momentum, will rapidly push $\nu_i \approx 0$ within machine precision even for very small coupling constants $\lambda$, rendering the proxy $\sim 1/\sqrt{\nu_i}$ useless.
Contrarily, in Fig.~\ref{fig:nu_evol} we see that a fast drop is stopped in DER after a few epochs.
The reason for this can be seen in Fig.~\ref{fig:ba_nu_evol} where we show the evolution of $\sqrt{\beta_i/\alpha_i}$ and $\sqrt{\nu_i / (1 + \nu_i)}$ which together are the width $w_\text{St}$ of the $t$-distribution,
\begin{equation}
    w_\text{St} = \sqrt{\frac{\beta_i (1 + \nu_i)}{\alpha_i \nu_i}}.
\end{equation}
It is straightforward to show that for large deviations, the gradients for $\gamma_i$ and $w_\text{St}$ push both towards larger values and are therefore aligned with decreasing values for $\nu_i$.
This is the second key insight to understand DER:
Optimizers are confused by the non-trivial correlation\footnote{Jacobian based optimizers cannot exactly follow curved gradients as induced by non-trivial correlations but approximately follow with piece-wise linear segments. As a consequence, optimizers overshoot and do not accumulate large momentum in these scenarios.} in $w_\text{St}$ with the parameters $\beta_i$ and $\alpha_i$ which effectively defers the minimization goal induced by the regularizer.
Once the gradient of $w_\text{St}$ has flipped sign, it depends on the coupling constant $\lambda$ how stiff $\nu_i$ is kept until the correct width, $w_\text{St} \approx 3$, is reached.
Only then, the optimizer starts to follow the \textit{valley} in the loss function (dashed line in Fig.~\ref{fig:ba_nu_evol}) that simultaneously preserves $w_\text{St}$ but decreases the total evidence.
Here, the residual is already small which limits the gradient induced by the regularizer.

\subsection{Redefining Proxies for Aleatoric and Epistemic Uncertainties}
\begin{figure*}
    \centering
    \begin{subfigure}[b]{.32\textwidth}
        \centering
        \includegraphics[width=\textwidth]{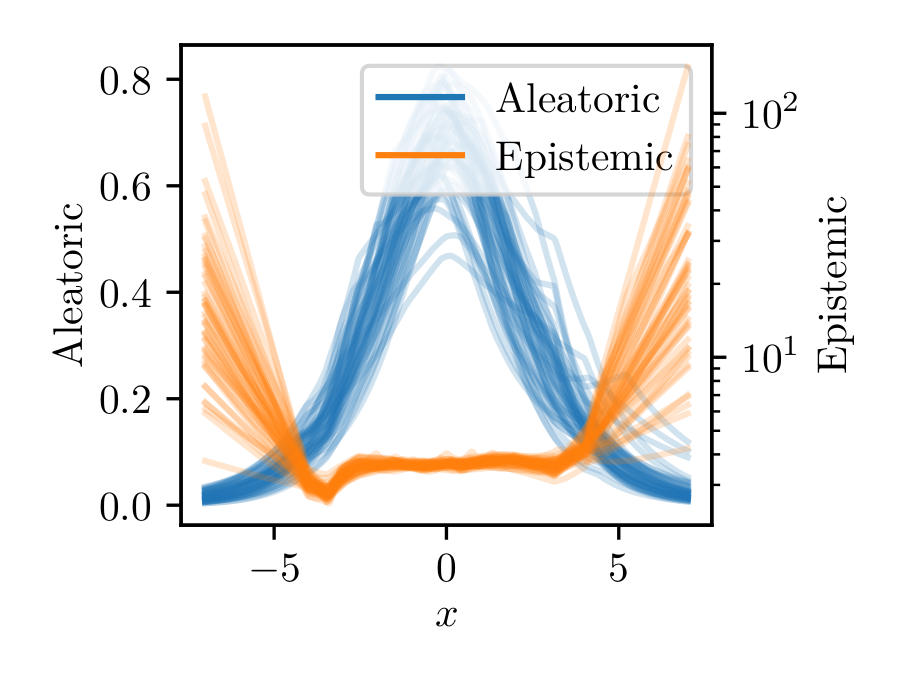}
        \caption{SOTA}
        \label{fig:uncertainties_sota}
    \end{subfigure}
    \hfill
    \begin{subfigure}[b]{.32\textwidth}
        \centering
        \includegraphics[width=\textwidth]{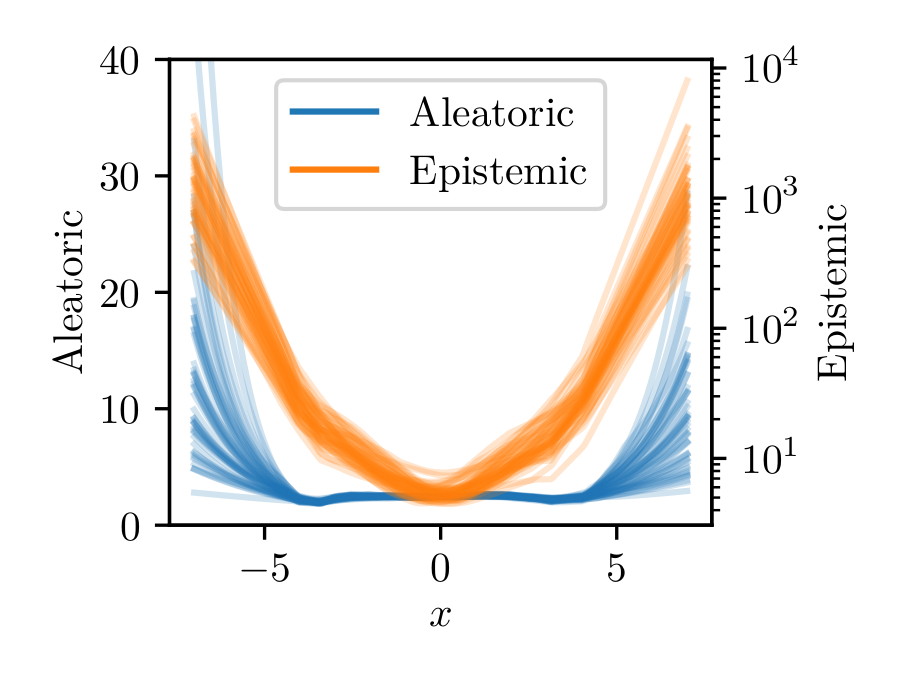}
        \caption{Proposed}
        \label{fig:uncertainties_new}
    \end{subfigure}
    \hfill
    \begin{subfigure}[b]{.32\textwidth}
        \centering
        \includegraphics[width=\textwidth]{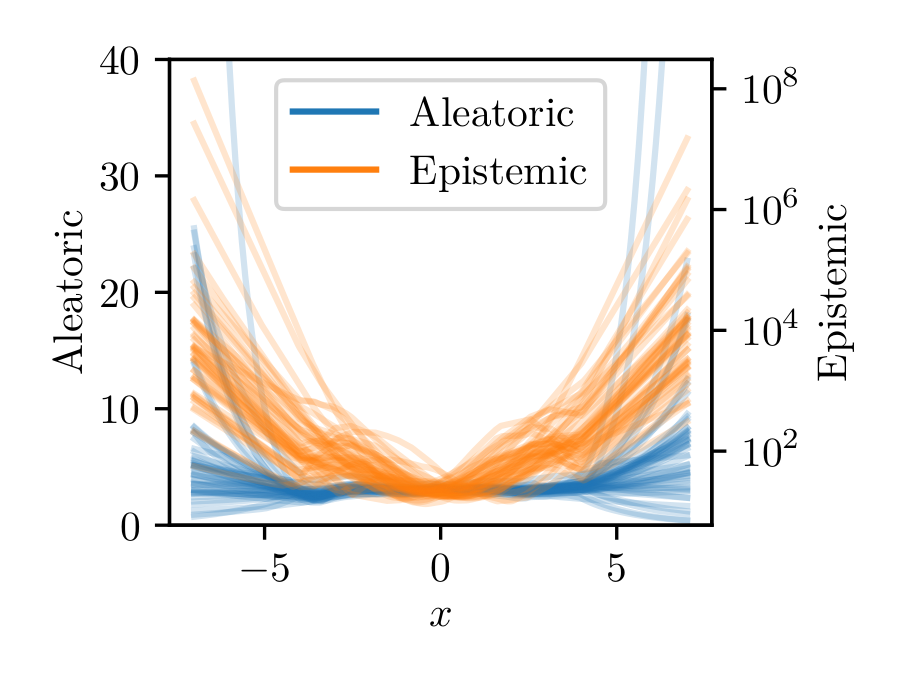}
        \caption{Alternative}
        \label{fig:uncertainties_new_gaussian}
    \end{subfigure}
    \caption{We propose a redefinition of the aleatoric and epistemic uncertainty. (Left) The prediction for both uncertainties is shown using the SOTA definitions after training the NN for 500 epochs. The training is repeated with $50$ different seeds. (Center) Results of the same NN but the uncertainties are estimated by our proposed definition in Eq.~\eqref{eq:uncertainties_new}. Note that most of the characteristic features of the epistemic uncertainty is shifted into the aleatoric uncertainty by our redefinition. (Right) Results of using the modified loss in Eq.~\eqref{eq:altloss}.}
    \label{fig:uncertainties}
\end{figure*}

We have shown above that the derivation of Eq.~\eqref{eq:loss} does not help to obtain the correct aleatoric or epistemic uncertainties in a strict Bayesian sense.
The fact that also the aleatoric prediction is spoiled follows directly by the insights we have described previously, but can also be seen visually by plotting $u_\text{al} = \sqrt{\mathbb{E}[\sigma_i^2]}$ as a function of $x$:
In Fig.~\ref{fig:uncertainties_sota} the prediction of the aleatoric uncertainty materializes as a peaking structure with a peak height of roughly $0.7$, whereas the data were generated with a constant standard deviation of $\sigma=3$ for all $x$.
Unsurprisingly and in accordance with Fig.~\ref{fig:ba_nu_evol}, this value is fitted in good approximation by the width of the $t$-distribution, $w_\text{St}$, due to the close resemblance between a $t$-distribution and a normal distribution.
Intuitively, the standard deviation of a normal distribution, approximated by the $w_\text{St}$, can be interpreted as the aleatoric uncertainty of the data and we therefore propose to redefine Eq.~\eqref{eq:uncertainties} accordingly.
Doing so unveils that the plateau between $-4 < x < +4$ of $\mathbb{E}[\sigma_i^2]$ was mainly a characteristic feature of the aleatoric and not the epistemic uncertainty!
In fact, it is reasonable to adopt the previous relation between aleatoric and epistemic uncertainty and redefine the latter as $1 / \sqrt{\nu_i}$ as shown in Fig.~\ref{fig:uncertainties_new}, which makes our proposed proxies for the aleatoric uncertainty $u_\text{al}'$ and the epistemic uncertainty $u_\text{ep}'$:
\begin{align}
    \label{eq:uncertainties_new}
    u_\text{al}' &\equiv w_\text{St} = \sqrt{\frac{\beta_i (1 + \nu_i)}{\alpha_i \nu_i}} &
    u_\text{ep}' &\equiv \frac{u_\text{ep}}{u_\text{al}} = \frac{1}{\sqrt{\nu_i}} \,.
\end{align}

Finally, we note that scaling the gradient of $\nu_i$ by the magnitude of the residual, regions with a high noise level also tend to contribute large gradients.
A possibility to disentangle this aleatoric contribution is to normalize the residual with $w_\text{St}$,
\begin{equation}
    \label{eq:newloss}
    \mathcal{L}_i(\boldsymbol{\omega}) = -\log L_i^\text{NIG}(\boldsymbol{\omega}) + \lambda \left| \frac{y_i - \gamma_i}{w_\text{St}} \right|^p \Phi.
\end{equation}
Most importantly, this inhibits the convergence for large, yet insignificant residuals in terms of the associated aleatoric uncertainty, in particular when the fit has reached the \textit{valley}.

For the synthetic example $(x_i, x_i^3 + \epsilon_i)$, we test the proposed change with $p=\{1,2\}$ and do not find any significant deviations w.r.t.\ the original formulation of the regularizer.
This is in good agreement with our expectation since in this example the variance of $\epsilon_i$ is constant over the generated sample.
However, in the Sec.~\ref{sec:binpulse} we test our proposed loss function on a data set with varying noise level and find that Eq.~\eqref{eq:newloss} more effectively strips this aleatoric component from the predicted epistemic uncertainty than DER with Eq.~\eqref{eq:loss}.

\subsection{Generalization and its Limitations}
In the Bayesian framework the parameter $\alpha_i$ is associated with the number of virtual measurements encoded in the prior.
In the limit $\alpha_i \to \infty$, the $t$-distribution with $2\alpha_i$ degrees of freedom becomes the normal distribution that was used for generating the synthetic data.
Na\"ively, one would therefore expect this number to be large at places with high model accuracy and data density, but, as we have eluded before, we find values $\alpha_i \approx 2$ instead.
In practice, this discrepancy causes a small offset between $w_\text{St}$ and the genuine standard deviation of the noise level.
Also, we find it noteworthy to mention that even if the data were genuinely distributed according to the $t$-distribution, the strong correlation of $\alpha_i$ and $w_\text{St}$ makes a clean extraction with a fitting approach challenging.

This observation motivates an extension of the mean-variance estimation by \citet{nix94} as a generalization of DER:\footnote{Unlike \citet{nix94}, we learn all parameters simultaneously.}
The likelihood of the $t$-distribution is replaced with a normal distribution $\mathcal{N}(y_i|\gamma_i, \sigma_i^2 = \beta_i / \nu_i)$ where $\beta_i$ is used to slow down the convergence of $\nu_i$.

Using our ansatz in Eq.~\eqref{eq:newloss} with $p=2$, the alternative loss functions reads
\begin{equation}
    \label{eq:altloss}
    \mathcal{L}'_i = \log \sigma_i^2 + (1 + \lambda \nu_i) \, \frac{(y_i - \gamma_i)^2}{\sigma_i^2}.
\end{equation}
Similarly, other data distributions, such as a Poisson or a Log-normal distribution, could be adopted by replacing the PDF of the normal distribution accordingly.
Clearly, this generalization only affects the aleatoric uncertainty estimation whereas the proxy for the epistemic uncertainty remains the same.
It is therefore pivotal to analyze, if the separation of the uncertainty types by stripping $\alpha_i$ from the equation preserves the good properties we have witnessed before.

We take our previous example\footnote{We also have used Eq.~\eqref{eq:altloss} for our example in Sec.~\ref{sec:binpulse}.} and use the modified loss function here.
We use the same NN, ignore $\theta_{i,3}$, and train it $50$ times with $\lambda=2$ and a learning rate of $5 \times 10^{-3}$ for 500 epochs.
The uncertainty predictions shown in Fig.~\ref{fig:uncertainties_new_gaussian} look similar, yet the sample variance is significantly higher.

This is what we finally coin the \textit{unreasonable effectiveness} of DER.
Defining the total evidence by two, rather than a single parameter, keeps the optimizer sufficiently busy during optimization and prevents a too fast convergence, thus reducing the dependency to the random initialization of $\nu_i$ and $\alpha_i$.
Furthermore, coupling $\alpha_i$ non-trivially in the NLL part of the loss function stabilizes the convergence even more.
This comes by the price of a bias due to the approximation of the standard deviation of the noise level with the width of a $t$-distribution.
For most applications the smaller sampling variance should offset this disadvantage, though.

\section{Experiments}
\subsection{Binary Pulse}
\label{sec:binpulse}
\begin{figure*}
    \centering
    \begin{subfigure}[b]{.32\textwidth}
        \centering
        \includegraphics[width=\linewidth]{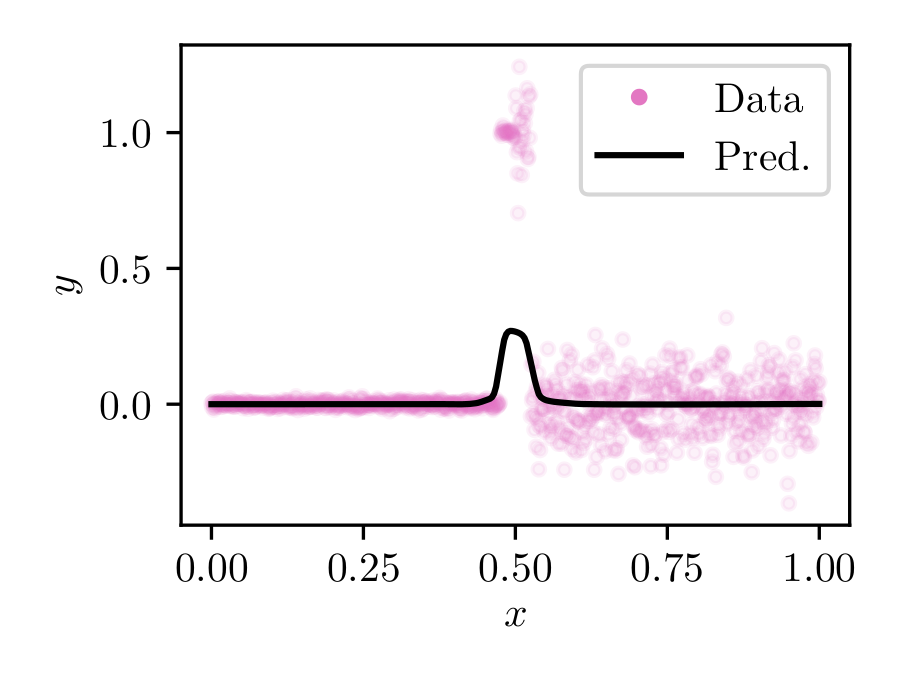}
        \caption{Data and prediction.} 
        \label{fig:binpulse}
    \end{subfigure}
    \hfill
    \begin{subfigure}[b]{.32\textwidth}
        \centering
        \includegraphics[width=\textwidth]{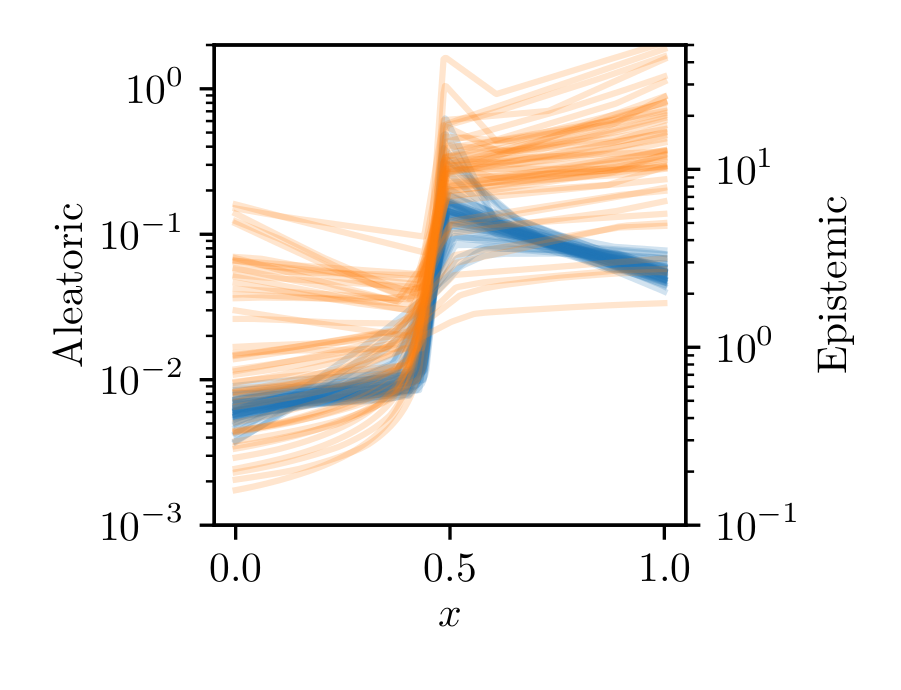}
        \caption{SOTA}
        \label{fig:uncertainties_new_binpulse_der}
    \end{subfigure}
    \hfill
    \begin{subfigure}[b]{.32\textwidth}
        \centering
        \includegraphics[width=\textwidth]{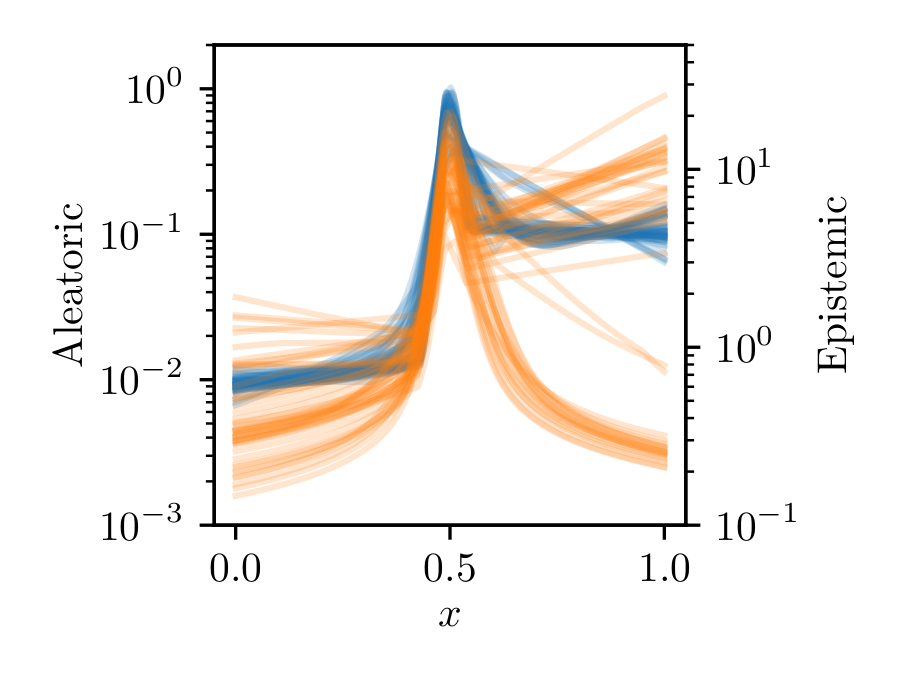}
        \caption{Proposed}
        \label{fig:uncertainties_new_binpulse_sder}
    \end{subfigure}
    \caption{Results of the binary pulse experiment. (Left) The averaged prediction $\gamma_i$ for our synthetic data sample as described in Eqs.~\eqref{eq:binpulse}. (Center) Estimations of the aleatoric (blue) and epistemic (orange) uncertainty using a regularizer proportional to the residual as used in the prior art. (Right) Uncertainty estimation using a normalized residual as proposed in Eq.~\eqref{eq:altloss}.}
\end{figure*}

The original formulation of DER uses the residual $|y_i - \gamma_i|$ to scale the gradients of the total evidence $\Phi$.
Apparently, this approach works fine if the variance of the noise level is constant for all training data.
If this is not the case, though, parts of the aleatoric uncertainty leak asymmetrically into the estimation of the epistemic uncertainty.
We demonstrate this with a simple synthetic data sample $(x_i, y_i + \epsilon_i)$ with a point-wise normal distributed noise $\epsilon_i \sim \mathcal{N}(0, \sigma_i^2)$ and
\begin{align}
    \label{eq:binpulse}
    y_i &= \begin{cases}
        1 & \text{if } |x_i - .5| < .0025 \\
        0 & \text{else}
    \end{cases} &
    \sigma_i^2 = \begin{cases}
        10^{-4} & \text{if } x_i < .5 \\
        10^{-2} & \text{else}.
    \end{cases}
\end{align}
We sample 1k data points and train the same shallow NN used previously, trained with $\lambda=0.01$ and a learning rate of $1 \times 10^{-3}$ for 600 epochs.
The data distribution and the corresponding predictions of the NN are shown in Fig.~\ref{fig:binpulse}.

The predicted uncertainties according to Eqs.~\eqref{eq:uncertainties_new} when trained with Eq.~\eqref{eq:loss} and Eq.~\eqref{eq:altloss} are shown in Fig.~\ref{fig:uncertainties_new_binpulse_der} and Fig.~\ref{fig:uncertainties_new_binpulse_sder}, respectively.

The synthetic data sample was chosen such that the model significantly under-fits the peak region and we expect to see a large epistemic uncertainty here.
Obviously, also the aleatoric uncertainty is pulled toward the peak in this region which is reasonable since large values of $\sigma_i^2$ indeed improve the total loss if the deviation is large.
More importantly, we see that with our modified version not only the aleatoric uncertainty prediction is better but also the epistemic uncertainty is symmetric, despite a certain sampling variance we have already witnessed before.

This example points towards a useful interpretation of $u_\text{al}'$ and $u_\text{ep}'$ from Eqs.~\eqref{eq:uncertainties_new}, in particular if paired with the modified loss~\eqref{eq:altloss}: The former is the point-wise prediction of the standard deviation of an additive, normal distributed noise source in the data, and the latter indicates when the estimation of the former is vague.

\subsection{Monocular Depth Estimation}
\begin{figure*}[t]
    \centering
    \begin{subfigure}[b]{.32\textwidth}
        \centering
        \includegraphics[width=\textwidth]{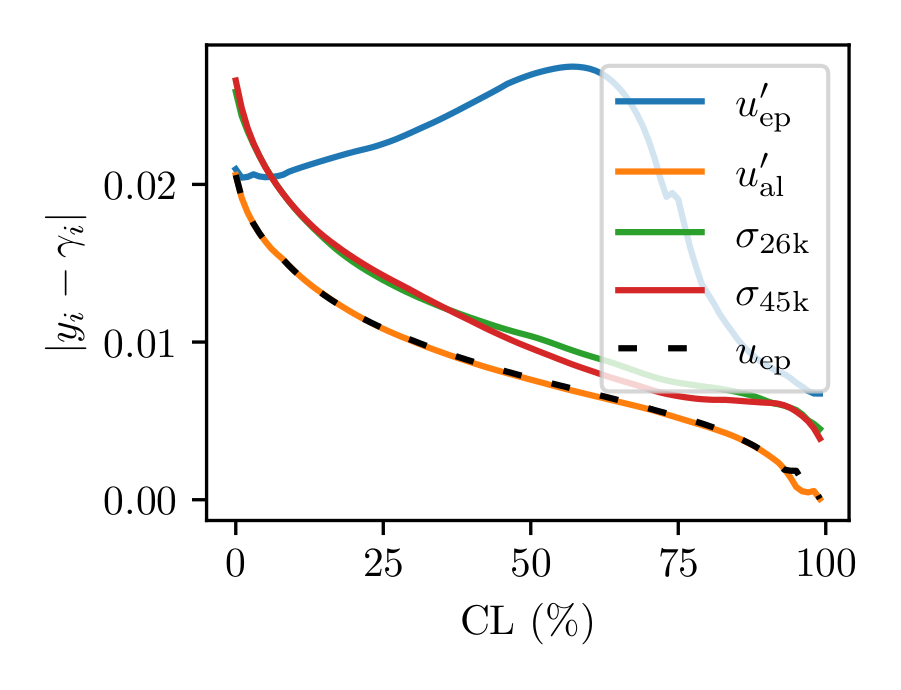}
        \caption{Prediction CL vs.\ observed error.}
        \label{fig:cutoffs}
    \end{subfigure}
    \hfill
    \begin{subfigure}[b]{.32\textwidth}
        \centering
        \includegraphics[width=\textwidth]{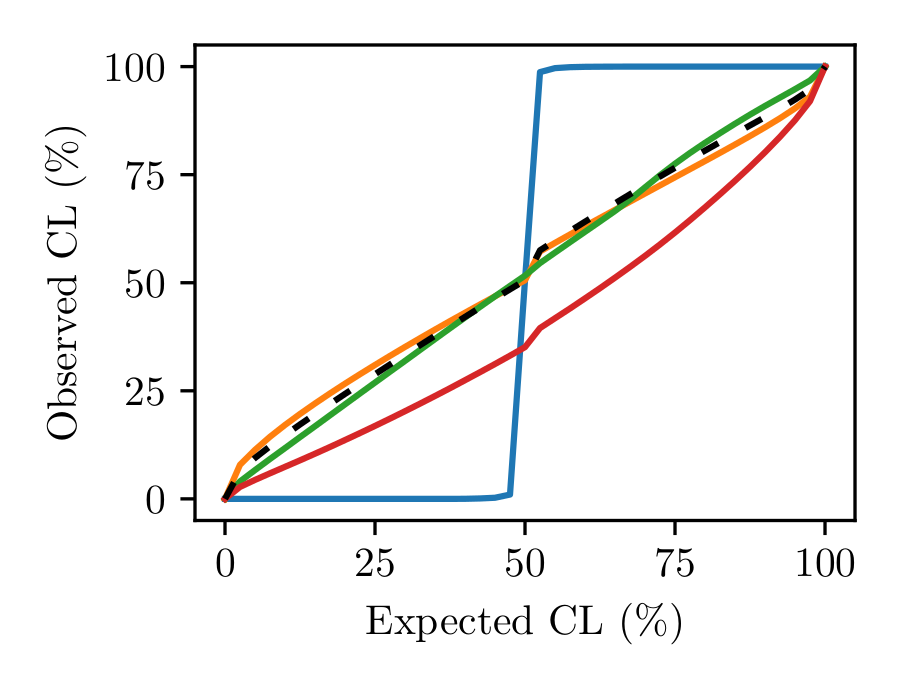}
        \caption{Uncertainty calibration.}
        \label{fig:calibs}
    \end{subfigure}
    \hfill
    \begin{subfigure}[b]{.32\textwidth}
        \centering
        \includegraphics[width=\linewidth]{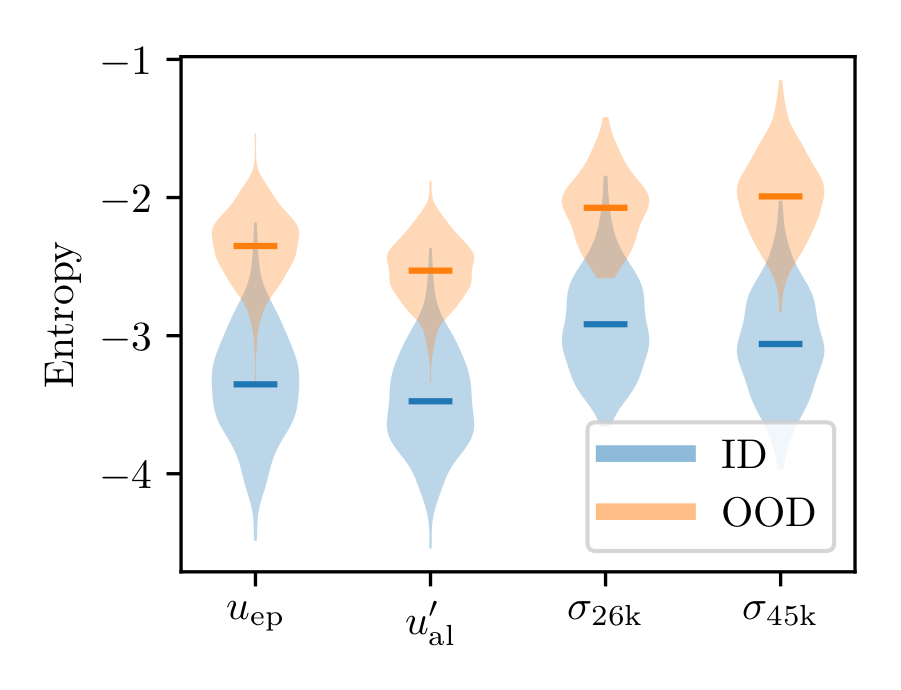}
        \caption{Violin plots of the entropy.} 
        \label{fig:entropy}
    \end{subfigure}
    \caption{(Left) Relationship between prediction CL and observed error. (Center) Model uncertainty calibration if $u_\text{ep}$ is replaced with $u_\text{ep}'$ or $u_\text{al}'$. A strong inverse trend for the former is desired. An ideal calibration is when the expected and the observed CL match. In addition, we also show the results of training a NLL of a normal distribution and using its standard deviation, $\sigma$, after 26k and 45k iterations. (Right) Violin plots of the entropy, $\log(2\pi \sigma^2) / 2$, where $\sigma=\{u_\text{ep}, u'_\text{al}, \sigma_\text{26k}, \sigma_\text{45k}\}$ is the respective prediction of the epistemic uncertainty. We compare the results of in-distribution data (ID) and out-of-distribution data (OOD) where the latter were not part of the training. The OOD data are taken from \citet{xinyu18}.}
    \label{fig:cutoffs_calibs}
\end{figure*}
Finally, we reevaluate the performance of DER on the same, high-dimensional task of depth estimation that was used in the prior art.
The training set consists of over 27k RGB-to-depth image pairs of indoor scenes from the NYU Depth v2 dataset \citep{silbermann12}.
We use the scripts\footnote{https://github.com/aamini/evidential-deep-learning, git commit hash: d1d8e39} for training and testing provided by \citet{amini20}, and only replace the actual estimation of $u_\text{ep}$ by applying the patches printed in the Appendix in Listing~\ref{lst:patch_ep} and Listing~\ref{lst:patch_al}.

Following \citet{kendall17,kuleshov18,amini20}, we show in Fig.~\ref{fig:cutoffs} how DER performs as pixels with uncertainty greater than certain thresholds are removed when the uncertainty is predicted by $u_\text{ep}$, $u_\text{ep}'$ or $u_\text{al}'$.
In full agreement with our previous findings, e.g., as shown in Fig.~\ref{fig:uncertainties}, our redefinition of the epistemic uncertainty, $u_\text{ep}$, shifts a significant part into the aleatoric uncertainty, $u_\text{al}'$.
The same behavior is seen in the calibration curves\footnote{For each predicted $\mu_i$ and $\sigma_i$, the inverse CDF of a normal distribution $\mathcal{N}(\mu_i, \sigma_i^2)$ is used to get the upper boundary of a confidence interval. The fraction of all predictions below this value is the observed CL. See the work of \citet{kendall17,kuleshov18} for more details.} shown in Fig.~\ref{fig:calibs}.
Our redefined proxy for the epistemic uncertainty, $u_\text{ep}'$, appears less helpful whereas $u_\text{ep} \approx u_\text{al}'$ in good approximation.

Since this is a complex data set we can only speculate about the reason for the apparent effectiveness of estimating epistemic uncertainty with a proxy for aleatoric uncertainty:
Still, in our previous experiments we already saw and described the phenomenon that the width of the $t$-distribution (or, similarly, the standard deviation of a normal distribution) tends to massively overshoot if the genuine model uncertainty is large.
In our experiments this overshooting was much larger than the genuine noise level of the data, hence, macroscopically and only if the genuine aleatoric uncertainty is sufficiently small, $u_\text{al}'$ indeed is a proxy for the epistemic uncertainty.

To verify our assumption we replace the loss function with Eq.~\eqref{eq:altloss} and set $\lambda = 0$, and obtain the canonical NLL of a normal distribution as discussed by \citet{nix94}.
The results of using the standard deviation of the normal distribution after 26k and 45k iterations as the proxy for the epistemic uncertainty are shown in Figs.~\ref{fig:cutoffs_calibs} and Fig.~\ref{fig:entropy}.
We observe a shift of a well-calibrated aleatoric proxy towards a less well calibrated, yet more powerful separator of in- and out-of-distribution data and anticipate that a similar transformation also leverages the good performance of DER in this task.

\section{Conclusion and Outlook}
Deep Evidential Regression (DER) is frequently used in the field as a tool to disentangle aleatoric and epistemic uncertainties.
We theoretically showed that the representation of the uncertainties is over parametrized making the efficiency of the DER seam almost unreasonable.
Using synthetic data, we demystified this unreasonable effectiveness of DER and related it to the convergence patterns of the learned uncertainty representations during the training process.
We also demonstrated that the estimation of the aleatoric uncertainty as used in the prior art yields quantitatively and qualitatively wrong results which brings us to the conclusion that $w_\text{St}$ does not disentangle aleatoric and epistemic uncertainties, however, it represents a reasonable proxy of the later in practice.

In our experiments with monocular depth estimation, we showed that the estimation of the epistemic uncertainty in the prior art \citep{amini20} is nearly equivalent to fitting the width of the $t$-distribution, $w_\text{St}$, a quantity that is, na\"ively, tightly coupled with the aleatoric uncertainty, missing the main disentanglement goal.

As a main result of this work we show that the behavior of DER can be understood with a fit of a Gaussian NLL, albeit, DER defers convergence speed such that $w_\text{St}$ becomes almost unreasonably effective.
Practitioners should treat DER more as a heuristic than an exact uncertainty quantification and carefully validate its predictions, while using the redefinitions and investigative studies we described here for disentangling uncertainties in NNs.

However, our results are negative in the following sense:
Although helpful in understanding the genuine meaning of DER predictions, we cannot fundamentally fix the underlying issue and deliver what was promised in the work of \citet{amini20}.
Instead, we like to motivate to investigate, given a specific application, whether it is worthwhile to strive for a precise epistemic uncertainty in the first place:
As pointed out by \citet{bengs22}, epistemic uncertainty is difficult to quantify objectively, because, unlike aleatoric uncertainty, epistemic uncertainty is not a property of the data or the data-generating process, and there is nothing like a \textit{ground truth} epistemic uncertainty.
If the consequence of a large epistemic uncertainty is to reject a model decision entirely and to fall back to, e.g., human supervision, the nominal value of the predicted uncertainty becomes less important and heuristics, such as DER, that do not require OOD data during training to supervise instances of high uncertainty, are actually still highly valuable.

More drastically, our findings of the Sec.~\ref{sec:sotaissue} that a projection cannot be used to unfold all degrees of freedom unambiguously might hint toward a similar fundamental problem that was reported recently by \citet{bengs22,huellermeier22}.
As a consequence, trying to exactly disentangle different types of uncertainties with \textit{second-order learners} with purely loss-based methods (e.g., evidential NNs) might be impossible and the Bayesian derivation of Eq.~\eqref{eq:studentt} is one way to prove it.

Finally, our observation of convergence speed as a proxy might circumvent the general problem with second-order learners and purely loss-based approaches that were studied by \citet{bengs22}.
Indeed, the convergence history can extract valuable information and, in a sense, does scale with the amount of information shown as labeled data during each epoch of a training sequence.
Hypothetically, our observation could be extended in future works to improve uncertainty prediction for regression but also for classification tasks and other related fields.

\section{Related Work}
Our work builds on the prior art \citep{amini20} for uncertainty estimation with evidential neural networks and the technical report of \citet{meinert21}, and more generally on the advancing area of uncertainty reasoning in deep learning.

In recent years there have been many explorations into Bayesian approaches to deep learning \citep{kendall17,neal96,guo17,wilson15,hafner19,ovadia19,izmailov19,seedat19}.
The key observation is that neural networks are typically underspecified by the data, thus different settings of the parameters correspond to a diverse variety of compelling explanations for the data --- i.e., a deep learning posterior consists of high performing models which make meaningfully different predictions on test data, as demonstrated by \citet{izmailov19,garipov18,zolna19}.
This underspecification by NNs makes Bayesian inference, and by corollary uncertainty estimation, particularly compelling for deep learning.
Bayesian deep learning aims to compute a distribution over the model parameters during training in order to quantify uncertainties,
such that the posterior is available for uncertainty estimation and model calibration \citep{guo17}.
With Bayesian NNs that have thousands and millions of parameters this posterior is intractable, so implementations largely focus on several approximate methods for Bayesian inference:
First, Markov Chain Monte Carlo (MCMC) methods
and in particular stochastic gradient MCMC for Bayesian NNs \citep{welling11,li16,park18,maddox19} show promise, with a main drawback being the inability to capture complex distributions in the parameter space without increasing the computational overhead.
Secondly, variational inference (VI) performs Bayesian inference by using a computationally tractable \textit{variational} distribution to approximate the posterior.
One approach by \citet{graves13} is to use a Gaussian variational posterior to approximate the distribution of the weights in a network, but the capacity of the uncertainty representation is limited by the variational distribution.
In general we see that MCMC has a higher variance and lower bias in the estimate, while VI has a higher bias but lower variance \citep{mattei19}.
The preeminent Bayesian deep learning approach by \citet{gal16} showed that variational inference can be approximated without modifying the network.
This is achieved through a method of approximate variational inference called Monte Carlo Dropout (MCD), whereby dropout is performed during inference, using multiple forward passes with randomly sampled dropout masks.
\citet{kendall17} used a combination of mean-variance estimation \citep{nix94} and MCD to simultaneously predict aleatoric and epistemic uncertainties.
However, this approach is limited by its requirement of multiple forward passes to gather enough information for a decent approximation by MCD which often makes this method uneconomical in practical applications.


Alternative to the prior-over-weights approach of Bayesian NN, one can view deep learning as an evidence acquisition process --- different from the Bayesian modeling nomenclature, evidence here is a measure of the amount of support collected from data in favor of a sample to be classified into a certain class, and uncertainty is inversely proportional to the total evidence \citep{sensoy18}.
Samples during training each add support to a learned higher-order, evidential distribution, which yields epistemic and aleatoric uncertainties without the need for sampling.
Several recent works develop this approach to deep learning and uncertainty estimation which put this in practice with \textit{prior networks} that place priors directly over the likelihood function \citep{amini20,malinin18}.
These approaches largely struggle with regularization \citep{sensoy18}, generalization (particularly without using out-of-distribution training data) \citep{malinin18,hafner19}, capturing aleatoric uncertainty \citep{gurevich19}, and the issues we have addressed above with the prior art Deep Evidential Regression \citep{amini20}.

There are also the frequentist approaches of bootstrapping and ensembling, which can be used to estimate NN uncertainty without the Bayesian computational overhead as well as being easily parallelizable --- for instance Deep Ensembles, where multiple randomly initialized NNs are trained and at test time the output variance from the ensemble of models is used as an estimate of uncertainty \citep{lakshminarayanan17}.

\section*{Reproducibility}
An open source GitHub repository with the source code for reproducing our experiments (implementations of the NNs and algorithms to generate the data) is available on https://github.com/pasteurlabs/unreasonable\_effective\_der. 
We encourage other researchers to reproduce, test, extend, and apply our work.
%
%

\bibliography{aaai23}

\begin{thebibliography}{40}
\providecommand{\natexlab}[1]{#1}

\bibitem[{Amini et~al.(2020)Amini, Schwarting, Soleimany, and Rus}]{amini20}
Amini, A.; Schwarting, W.; Soleimany, A.; and Rus, D. 2020.
\newblock {D}eep {E}vidential {R}egression.
\newblock \emph{Advances in Neural Information Processing Systems}, 33.

\bibitem[{Bengs, H\"ullermeier, and Waegeman(2022)}]{bengs22}
Bengs, V.; H\"ullermeier, E.; and Waegeman, W. 2022.
\newblock Pitfalls of Epistemic Uncertainty Quantification through Loss
  Minimisation.
\newblock \emph{Advances in Neural Information Processing Systems}, 35.

\bibitem[{Cai et~al.(2021)Cai, Wang, Huang, Liu, and Liu}]{cai21}
Cai, P.; Wang, H.; Huang, H.; Liu, Y.; and Liu, M. 2021.
\newblock Vision-Based Autonomous Car Racing Using {D}eep {I}mitative
  {R}einforcement {L}earning.
\newblock \emph{IEEE Robotics and Automation Letters}, 6(4).

\bibitem[{Chen, Bromuri, and van Eekelen(2021)}]{chen21}
Chen, X.; Bromuri, S.; and van Eekelen, M. 2021.
\newblock \emph{{N}eural Machine Translation for Harmonized System Codes
  Prediction}.
\newblock Association for Computing Machinery.
\newblock ISBN 978-1-450-38940-2.

\bibitem[{Gal and Ghahramani(2016)}]{gal16}
Gal, Y.; and Ghahramani, Z. 2016.
\newblock Dropout as a {B}ayesian Approximation: Representing Model Uncertainty
  in {D}eep {L}earning.
\newblock \emph{International Conference on Machine Learning}, 48.

\bibitem[{Garipov et~al.(2018)Garipov, Izmailov, Podoprikhin, Vetrov, and
  Wilson}]{garipov18}
Garipov, T.; Izmailov, P.; Podoprikhin, D.; Vetrov, D.; and Wilson, A.~G. 2018.
\newblock Loss Surfaces, Mode Connectivity, and Fast Ensembling of {DNN}s.
\newblock \emph{Advances in Neural Information Processing Systems}, 31.

\bibitem[{Gelman et~al.(2013)Gelman, Carlin, Stern, Dunson, Vehtari, and
  Rubin}]{gelman13}
Gelman, A.; Carlin, J.; Stern, H.; Dunson, D.; Vehtari, A.; and Rubin, D. 2013.
\newblock \emph{{B}ayesian Data Analysis}.
\newblock Taylor \& Francis Ltd.
\newblock ISBN 978-1-439-84095-5.

\bibitem[{Graves, Mohamed, and Hinton(2013)}]{graves13}
Graves, A.; Mohamed, A.; and Hinton, G. 2013.
\newblock Speech Recognition with {D}eep {R}ecurrent {N}eural {N}etworks.
\newblock \emph{IEEE International Conference on Acoustics, Speech, and Signal
  Processing}.

\bibitem[{Guo et~al.(2017)Guo, Pleiss, Sun, and Weinberger}]{guo17}
Guo, C.; Pleiss, G.; Sun, Y.; and Weinberger, K.~Q. 2017.
\newblock On Calibration of Modern {N}eural {N}etworks.
\newblock \emph{International Conference on Machine Learning}, 34.

\bibitem[{Gurevich and Stuke(2019)}]{gurevich19}
Gurevich, P.; and Stuke, H. 2019.
\newblock Gradient conjugate priors and multi-layer {N}eural {N}etworks.
\newblock \emph{Artificial Intelligence}, 278.

\bibitem[{Hafner et~al.(2019)Hafner, Tran, Lillicrap, Irpan, and
  Davidson}]{hafner19}
Hafner, D.; Tran, D.; Lillicrap, T.; Irpan, A.; and Davidson, J. 2019.
\newblock Noise Contrastive Priors for Functional Uncertainty.
\newblock \emph{Proceedings of Machine Learning Research}, 115.

\bibitem[{Hora(1996)}]{hora96}
Hora, S.~C. 1996.
\newblock Aleatory and epistemic uncertainty in probability elicitation with an
  example from hazardous waste management.
\newblock \emph{Reliability Engineering \& System Safety}, 54(2-3).

\bibitem[{Huang et~al.(2018)Huang, Cheng, Geng, Cao, Zhou, Wang, Lin, and
  Yang}]{xinyu18}
Huang, X.; Cheng, X.; Geng, Q.; Cao, B.; Zhou, D.; Wang, P.; Lin, Y.; and Yang,
  R. 2018.
\newblock The {A}polloScape Dataset for Autonomous Driving.
\newblock \emph{IEEE/CVF Conference on Computer Vision and Pattern Recognition
  Workshops}.

\bibitem[{H\"ullermeier(2022)}]{huellermeier22}
H\"ullermeier, E. 2022.
\newblock Quantifying Aleatoric and Epistemic Uncertainty in Machine Learning:
  Are Conditional Entropy and Mutual Information Appropriate Measures?
\newblock \emph{arXiv:2209.03302 [cs.LG]}.

\bibitem[{H\"ullermeier and Waegeman(2021)}]{huellermeier21}
H\"ullermeier, E.; and Waegeman, W. 2021.
\newblock Aleatoric and epistemic uncertainty in machine learning: an
  introduction to concepts and methods.
\newblock \emph{Machine Learning}, 110.

\bibitem[{Izmailov et~al.(2019)Izmailov, Maddox, Kirichenko, Garipov, Vetrov,
  and Wilson}]{izmailov19}
Izmailov, P.; Maddox, W.~J.; Kirichenko, P.; Garipov, T.; Vetrov, D.~P.; and
  Wilson, A.~G. 2019.
\newblock Subspace Inference for {B}ayesian {D}eep {L}earning.
\newblock \emph{Proceedings of Machine Learning Research}, 115.

\bibitem[{Kendall and Gal(2017)}]{kendall17}
Kendall, A.; and Gal, Y. 2017.
\newblock What Uncertainties Do We Need in {B}ayesian {D}eep {L}earning for
  {C}omputer {V}ision?
\newblock \emph{Advances in Neural Information Processing Systems}, 30.

\bibitem[{Kiureghian and Ditlevsen(2009)}]{kiureghian09}
Kiureghian, A.~D.; and Ditlevsen, O. 2009.
\newblock Aleatory or epistemic? Does it matter?
\newblock \emph{Structural Safety}, 31(2).

\bibitem[{Kuleshov, Fenner, and Ermon(2018)}]{kuleshov18}
Kuleshov, V.; Fenner, N.; and Ermon, S. 2018.
\newblock Accurate Uncertainties for {D}eep {L}earning Using Calibrated
  Regression.
\newblock \emph{International Conference on Machine Learning}, 35.

\bibitem[{Lakshminarayanan, Pritzel, and Blundell(2017)}]{lakshminarayanan17}
Lakshminarayanan, B.; Pritzel, A.; and Blundell, C. 2017.
\newblock Simple and Scalable Predictive Uncertainty Estimation using {D}eep
  {E}nsembles.
\newblock \emph{Advances in Neural Information Processing Systems}, 30.

\bibitem[{Li et~al.(2016)Li, Stevens, Chen, Pu, Gan, and Carin}]{li16}
Li, C.; Stevens, A.; Chen, C.; Pu, Y.; Gan, Z.; and Carin, L. 2016.
\newblock Learning Weight Uncertainty with Stochastic Gradient {MCMC} for Shape
  Classification.
\newblock \emph{IEEE Conference on Computer Vision and Pattern Recognition}.

\bibitem[{Li and Liu(2022)}]{li22}
Li, H.; and Liu, J. 2022.
\newblock 3D High-Quality Magnetic Resonance Image Restoration in Clinics Using
  {D}eep {L}earning.

\bibitem[{Liu et~al.(2021)Liu, Amini, Zhu, Karaman, Han, and Rus}]{zhijian21}
Liu, Z.; Amini, A.; Zhu, S.; Karaman, S.; Han, S.; and Rus, D. 2021.
\newblock Efficient and Robust {LiDAR}-Based End-to-End Navigation.
\newblock \emph{IEEE International Conference on Robotics and Automation}.

\bibitem[{Maddox et~al.(2019)Maddox, Garipov, Izmailov, Vetrov, and
  Wilson}]{maddox19}
Maddox, W.; Garipov, T.; Izmailov, P.; Vetrov, D.; and Wilson, A.~G. 2019.
\newblock A Simple Baseline for {B}ayesian Uncertainty in {D}eep {L}earning.
\newblock \emph{Advances in Neural Information Processing Systems}, 32.

\bibitem[{Malinin and Gales(2018)}]{malinin18}
Malinin, A.; and Gales, M. 2018.
\newblock Predictive Uncertainty Estimation via {P}rior {N}etworks.
\newblock \emph{Advances in Neural Information Processing Systems}, 31.

\bibitem[{Mattei(2020)}]{mattei19}
Mattei, P.-A. 2020.
\newblock A Parsimonious Tour of {B}ayesian Model Uncertainty.
\newblock \emph{arXiv:1902.05539 [stat.ME]}.

\bibitem[{Meinert and Lavin(2021)}]{meinert21}
Meinert, N.; and Lavin, A. 2021.
\newblock Multivariate {D}eep {E}vidential {R}egression.
\newblock \emph{arXiv:2104.06135 [cs.LG]}.

\bibitem[{Neal(1996)}]{neal96}
Neal, R.~M. 1996.
\newblock \emph{{B}ayesian Learning for {N}eural {N}etworks}.
\newblock Springer.
\newblock ISBN 978-1-4612-0745-0.

\bibitem[{Nix and Weigend(1994)}]{nix94}
Nix, D.~A.; and Weigend, A.~S. 1994.
\newblock Estimating the mean and variance of the target probability
  distribution.
\newblock \emph{IEEE International Conference on Neural Networks}.

\bibitem[{Ovadia et~al.(2010)Ovadia, Fertig, Ren, Nado, Sculley, Nowozin,
  Dillon, Lakshminarayanan, and Snoek}]{ovadia19}
Ovadia, Y.; Fertig, E.; Ren, J.; Nado, Z.; Sculley, D.; Nowozin, S.; Dillon,
  J.~V.; Lakshminarayanan, B.; and Snoek, J. 2010.
\newblock Can You Trust Your Model's Uncertainty? Evaluating Predictive
  Uncertainty Under Dataset Shift.
\newblock \emph{Advances in Neural Information Processing Systems}, 32.

\bibitem[{Park et~al.(2018)Park, Kim, Ha, and Lee}]{park18}
Park, C.; Kim, J.~M.; Ha, S.~H.; and Lee, J. 2018.
\newblock Sampling-based {B}ayesian Inference with gradient uncertainty.
\newblock \emph{Workshop on Bayesian Deep Learning}.

\bibitem[{Petek et~al.(2022)Petek, Sirohi, Büscher, and Burgard}]{petek22}
Petek, K.; Sirohi, K.; Büscher, D.; and Burgard, W. 2022.
\newblock Robust Monocular Localization in Sparse {HD} Maps Leveraging
  Multi-Task Uncertainty Estimation.

\bibitem[{Seedat and Kanan(2020)}]{seedat19}
Seedat, N.; and Kanan, C. 2020.
\newblock Towards calibrated and scalable uncertainty representations for
  {N}eural {N}etworks.
\newblock \emph{Advances in Neural Information Processing Systems}, 33.

\bibitem[{Sensoy, Kaplan, and Kandemir(2018)}]{sensoy18}
Sensoy, M.; Kaplan, L.; and Kandemir, M. 2018.
\newblock {E}vidential {D}eep {L}earning to Quantify Classification
  Uncertainty.
\newblock \emph{Advances in Neural Information Processing Systems}, 31.

\bibitem[{Silberman et~al.(2012)Silberman, Hoiem, Kohli, and
  Fergus}]{silbermann12}
Silberman, N.; Hoiem, D.; Kohli, P.; and Fergus, R. 2012.
\newblock Indoor Segmentation and Support Inference from {RGBD} Images.
\newblock \emph{European Conference on Computer Vision}.

\bibitem[{Singh et~al.(2022)Singh, Fowdur, Gawlikowski, and Medina}]{singh22}
Singh, S.~K.; Fowdur, J.~S.; Gawlikowski, J.; and Medina, D. 2022.
\newblock Leveraging {E}vidential {D}eep {L}earning Uncertainties with
  Graph-based Clustering to Detect Anomalies.
\newblock \emph{IEEE Transactions on Intelligent Transportation Systems}, 25.

\bibitem[{Soleimany et~al.(2021)Soleimany, Amini, Goldman, Rus, Bhatia, and
  Coley}]{soleimany21}
Soleimany, A.~P.; Amini, A.; Goldman, S.; Rus, D.; Bhatia, S.~N.; and Coley,
  C.~W. 2021.
\newblock {E}vidential {D}eep {L}earning for Guided Molecular Property
  Prediction and Discovery.
\newblock \emph{ACS Central Science}, 7(8).

\bibitem[{Welling and Teh(2011)}]{welling11}
Welling, M.; and Teh, Y.~W. 2011.
\newblock {B}ayesian learning via stochastic gradient langevin dynamics.
\newblock \emph{International Conference on Machine Learning}, 28.

\bibitem[{Wilson et~al.(2016)Wilson, Hu, Salakhutdinov, and Xing}]{wilson15}
Wilson, A.~G.; Hu, Z.; Salakhutdinov, R.; and Xing, E.~P. 2016.
\newblock {D}eep {K}ernel Learning.
\newblock \emph{International Conference on Artificial Intelligence and
  Statistics}, 19.

\bibitem[{Zolna, Geras, and Cho(2020)}]{zolna19}
Zolna, K.; Geras, K.~J.; and Cho, K. 2020.
\newblock Classifier-agnostic saliency map extraction.
\newblock \emph{Computer Vision and Image Understanding}, 196.

\end{thebibliography}

\appendix

\begin{figure*}[ht]
    \centering
    \begin{subfigure}[b]{.33\textwidth}
        \centering
        \includegraphics[width=\textwidth]{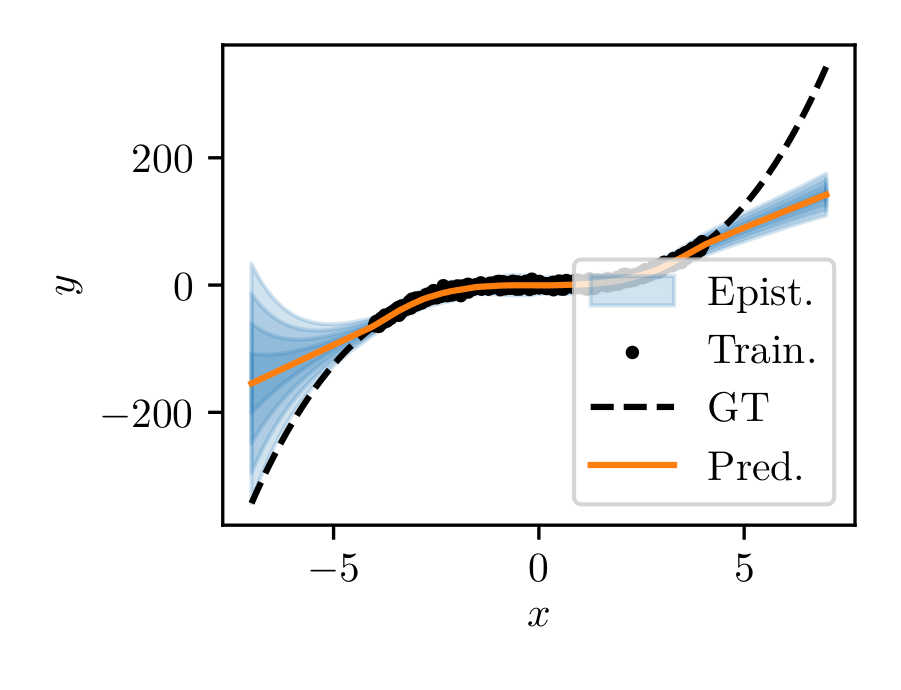}
    \end{subfigure}
    \hfill
    \begin{subfigure}[b]{.33\textwidth}
        \centering
        \includegraphics[width=\textwidth]{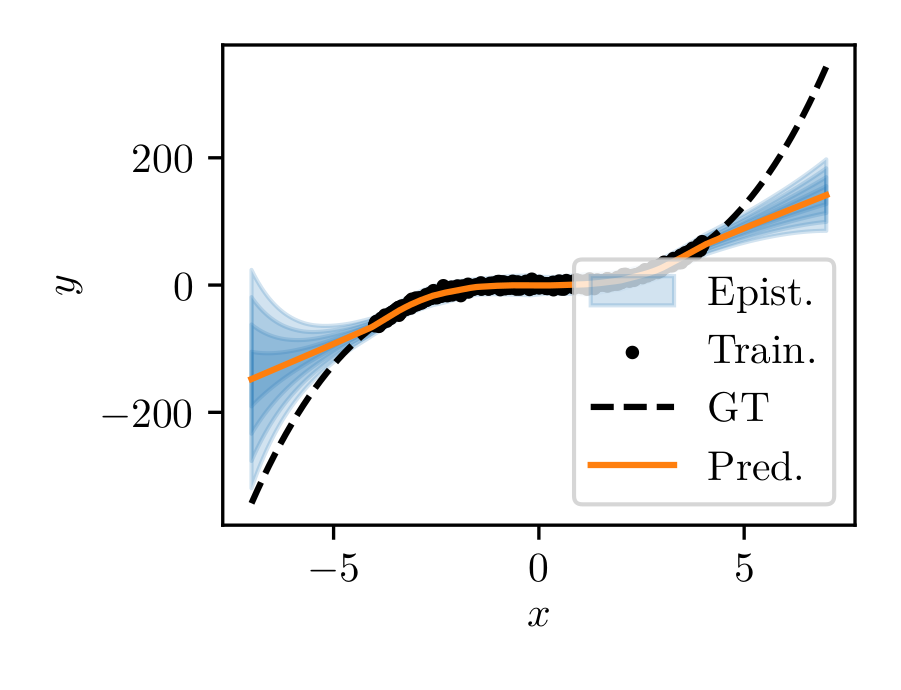}
    \end{subfigure}
    \hfill
    \begin{subfigure}[b]{.33\textwidth}
        \centering
        \includegraphics[width=\textwidth]{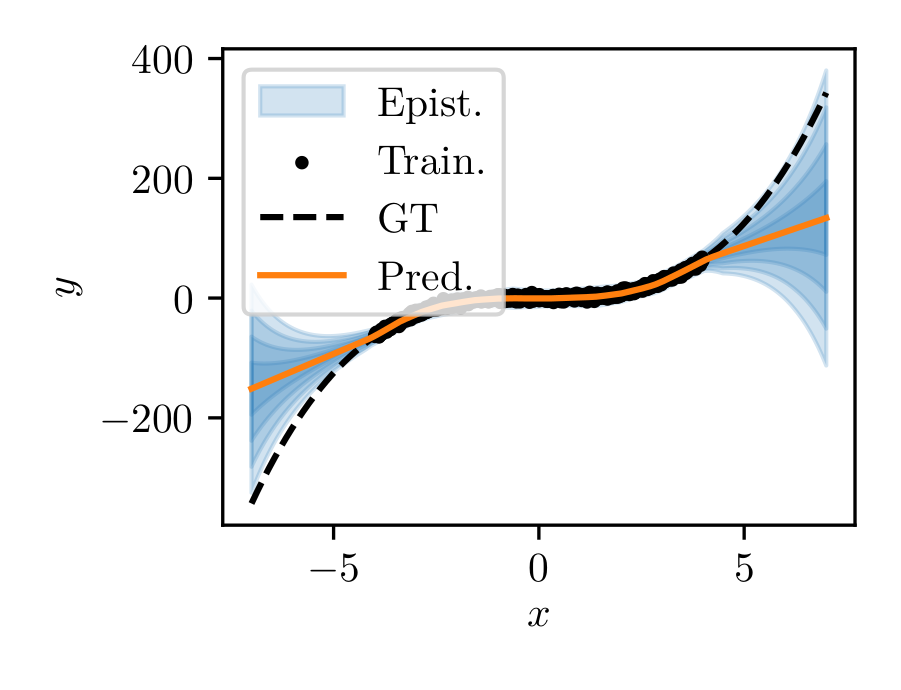}
    \end{subfigure} \\
    \begin{subfigure}[b]{.33\textwidth}
        \centering
        \includegraphics[width=\textwidth]{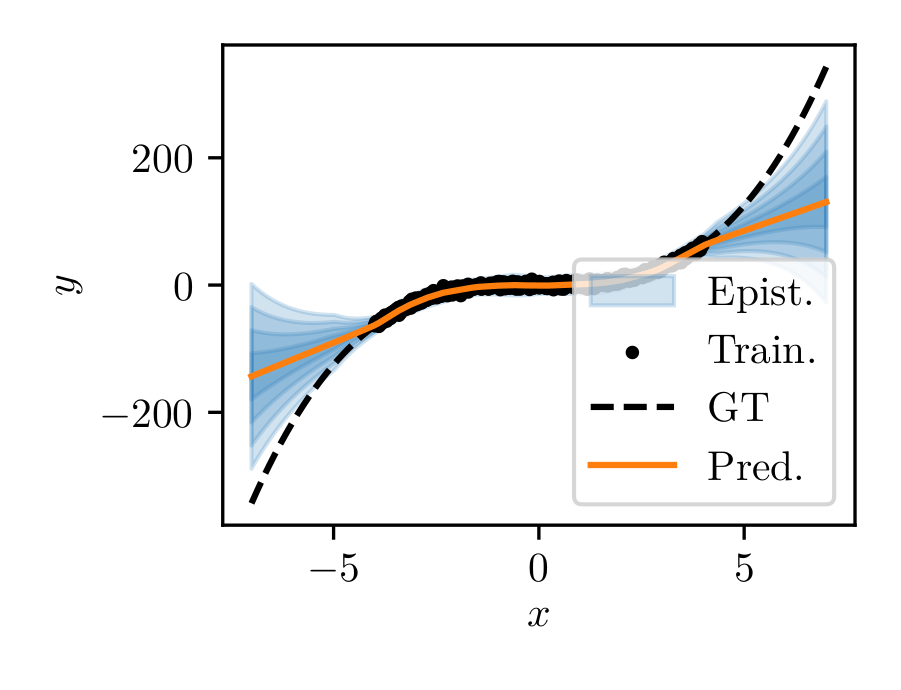}
    \end{subfigure}
    \hfill
    \begin{subfigure}[b]{.33\textwidth}
        \centering
        \includegraphics[width=\textwidth]{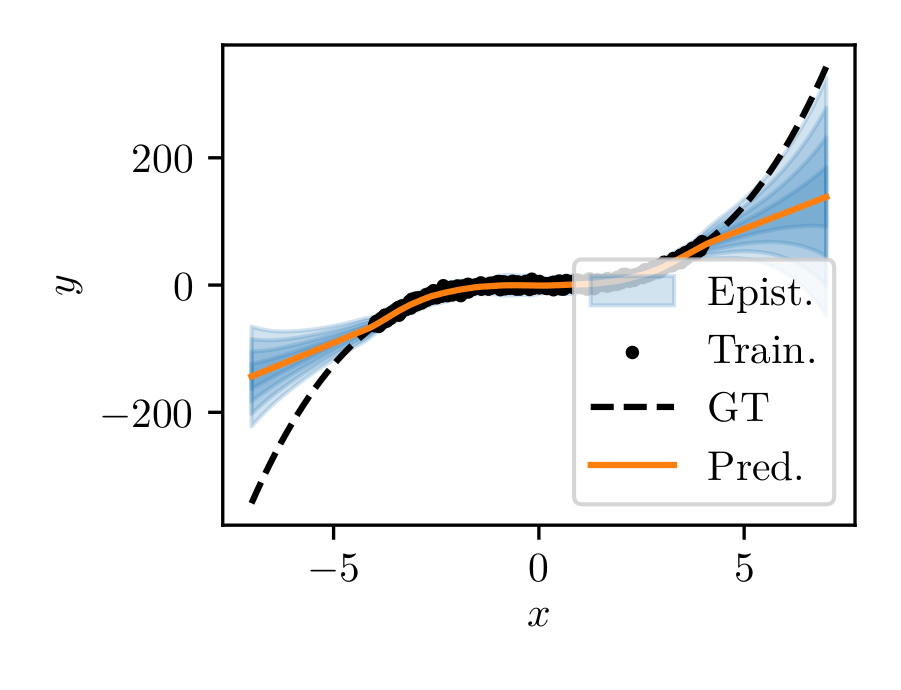}
    \end{subfigure}
    \hfill
    \begin{subfigure}[b]{.33\textwidth}
        \centering
        \includegraphics[width=\textwidth]{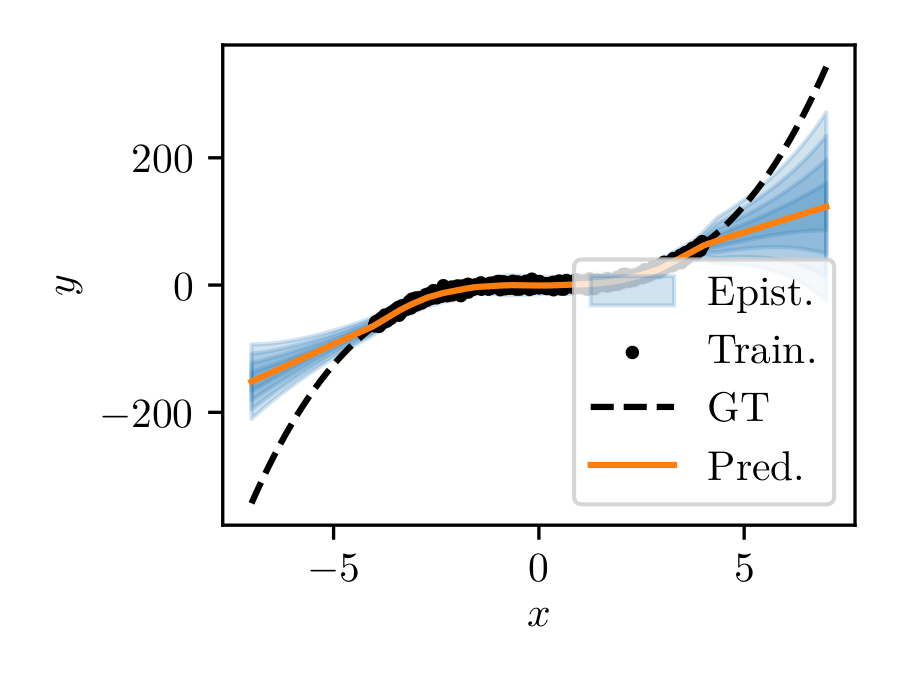}
    \end{subfigure} \\
        \begin{subfigure}[b]{.33\textwidth}
        \centering
        \includegraphics[width=\textwidth]{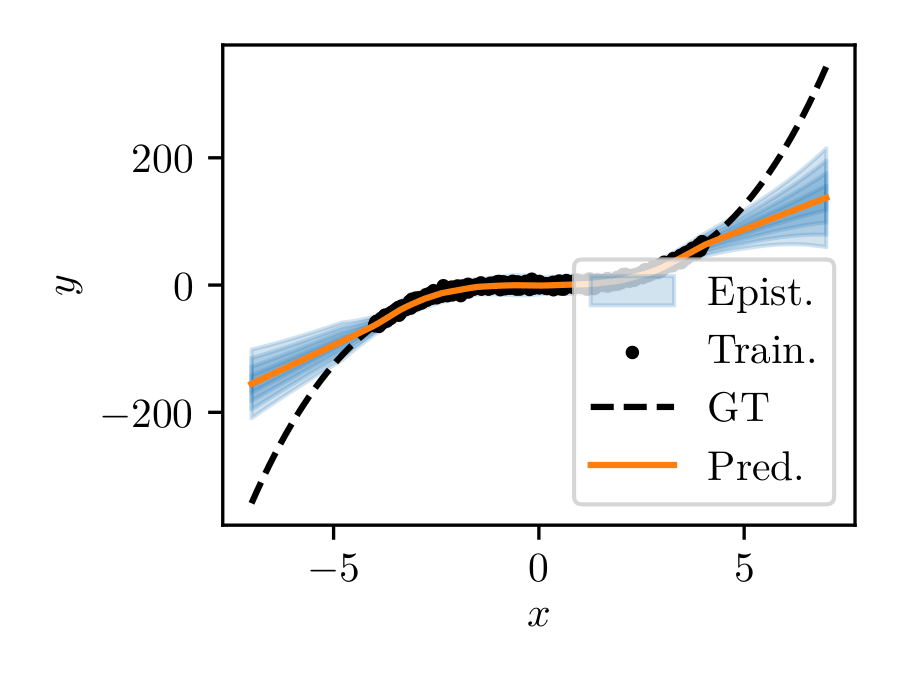}
    \end{subfigure}
    \hfill
    \begin{subfigure}[b]{.33\textwidth}
        \centering
        \includegraphics[width=\textwidth]{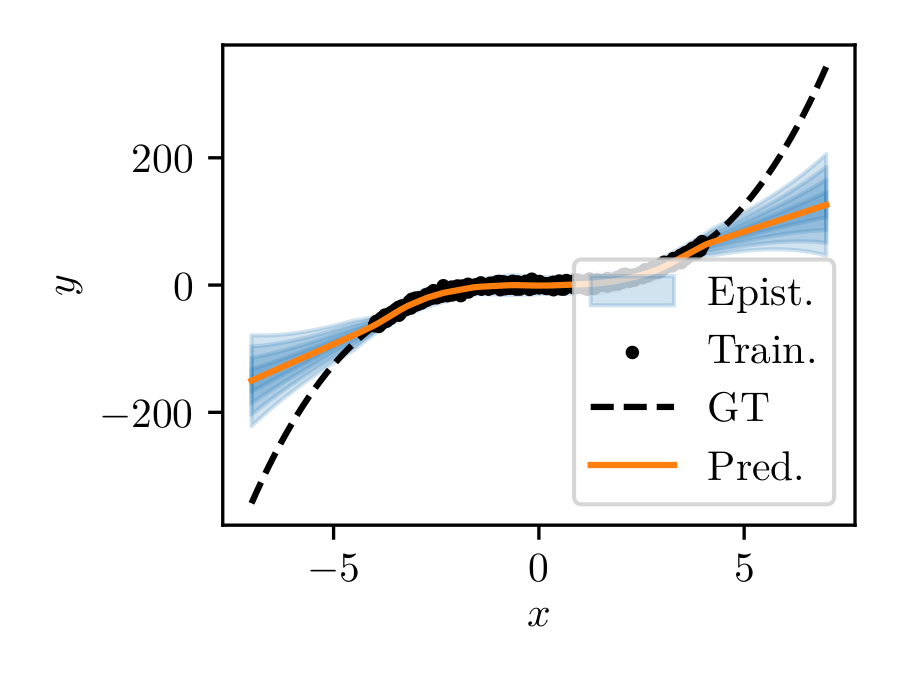}
    \end{subfigure}
    \hfill
    \begin{subfigure}[b]{.33\textwidth}
        \centering
        \includegraphics[width=\textwidth]{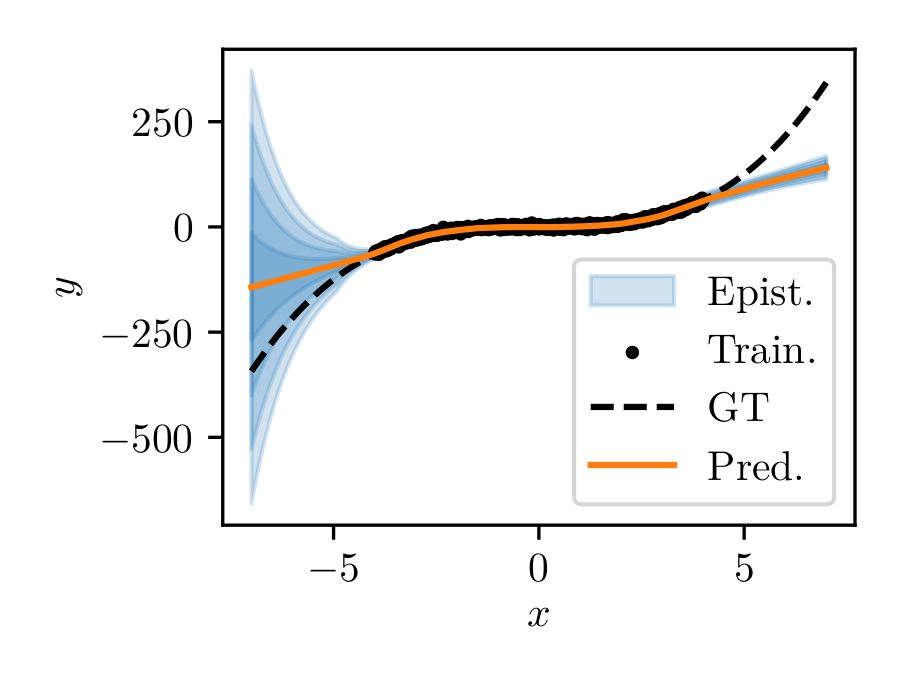}
    \end{subfigure}
    \caption{Results of reproducing the original synthetic experiment by \citet{amini20}. We train $50$ model individually on 1k data points each. The result of the first nine models are shown.}
    \label{fig:sota}
\end{figure*}

\begin{figure*}[ht]
    \centering
    \begin{subfigure}[b]{.45\textwidth}
        \centering
        \includegraphics[width=\textwidth]{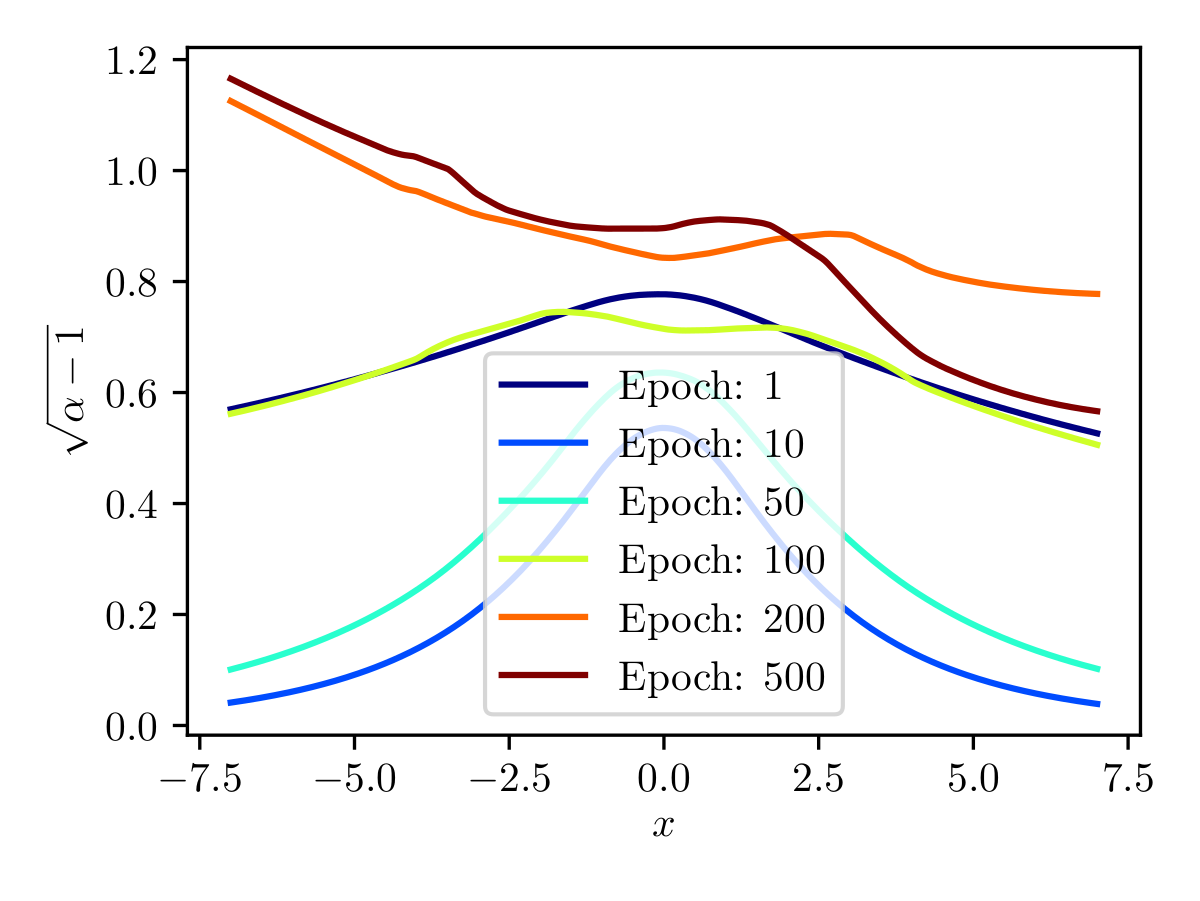}
        \caption{Evolution of $\alpha$.}
        \label{fig:alpha_evol}
    \end{subfigure}
    \hfill
    \begin{subfigure}[b]{.45\textwidth}
        \centering
        \includegraphics[width=\textwidth]{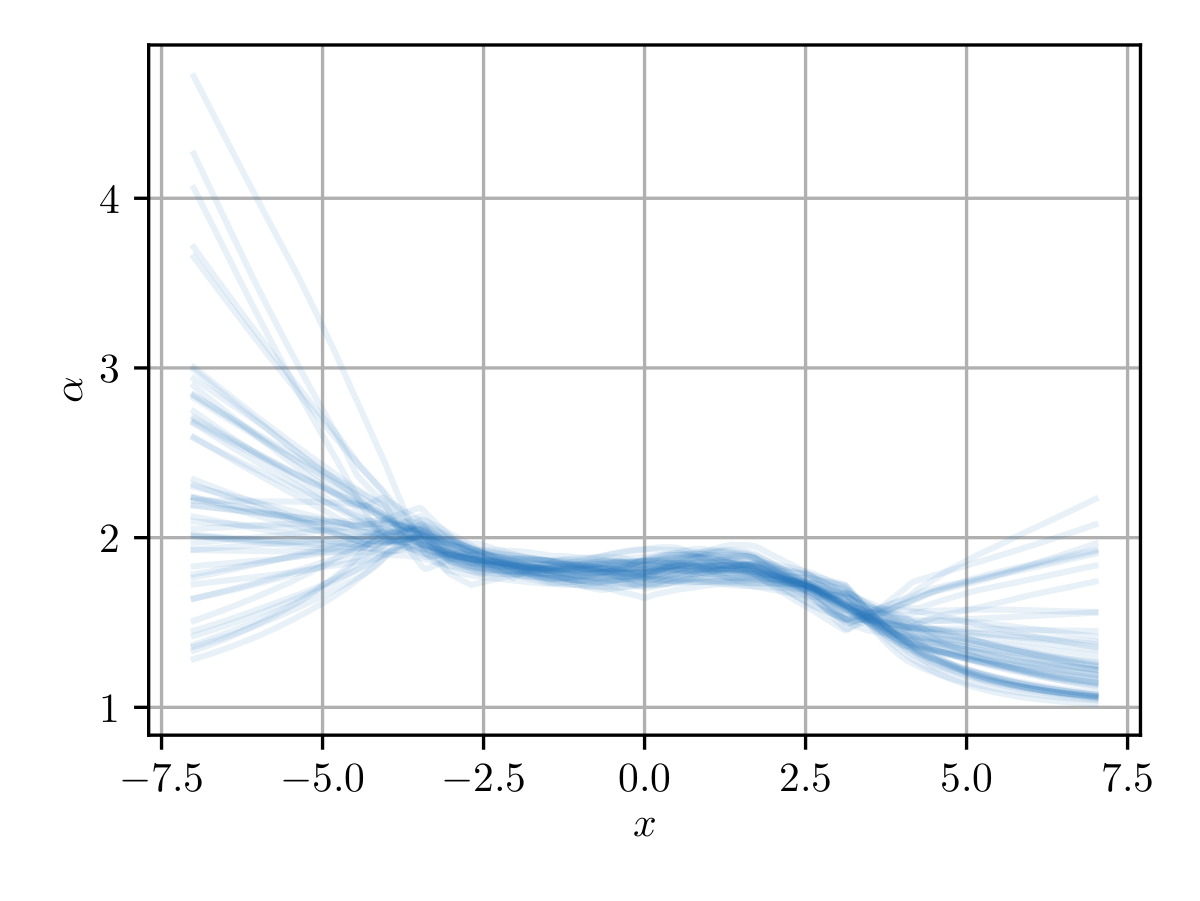}
        \caption{Distribution of $\alpha$ in last epoch.}
        \label{fig:alpha}
    \end{subfigure} \\
    \begin{subfigure}[b]{.45\textwidth}
        \centering
        \includegraphics[width=\textwidth]{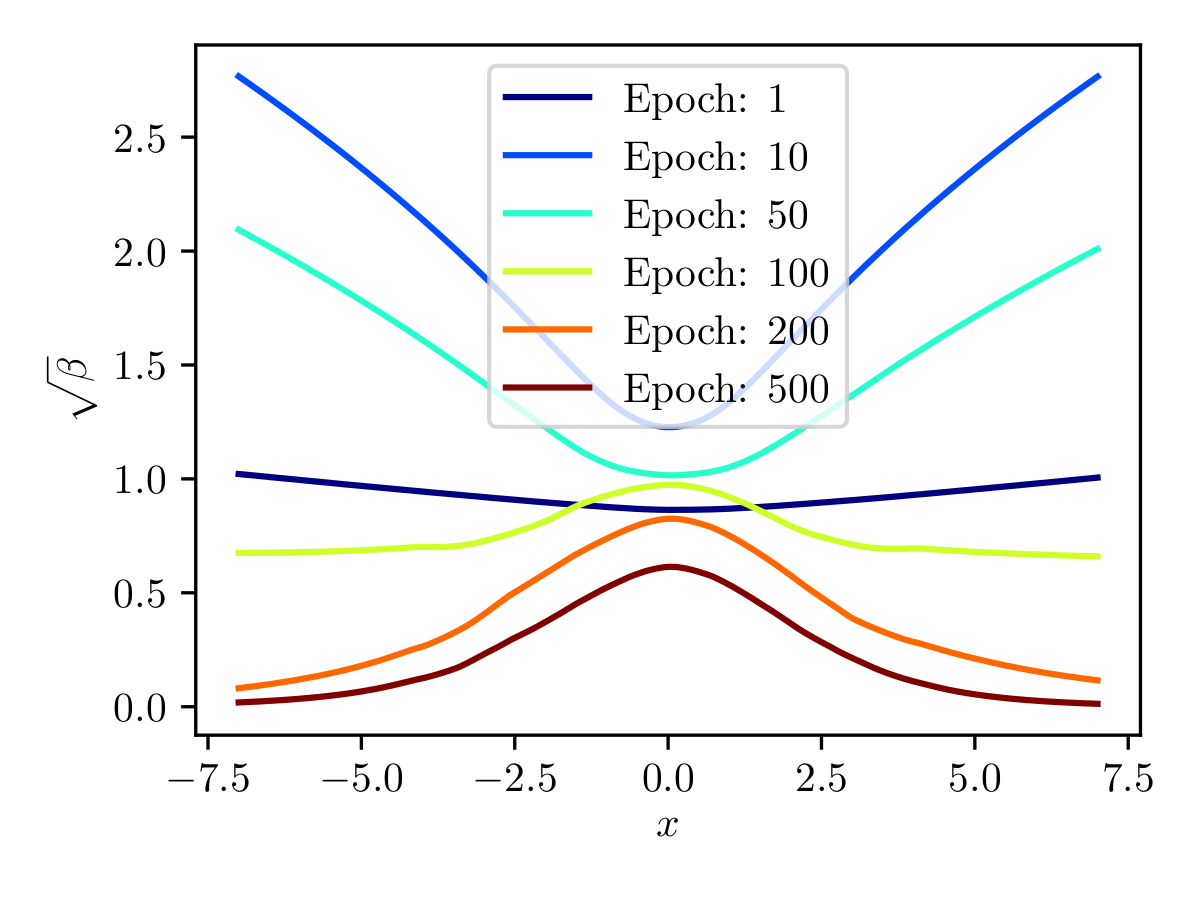}
        \caption{Evolution of $\beta$.}
        \label{fig:beta_evol}
    \end{subfigure}
    \hfill
    \begin{subfigure}[b]{.45\textwidth}
        \centering
        \includegraphics[width=\textwidth]{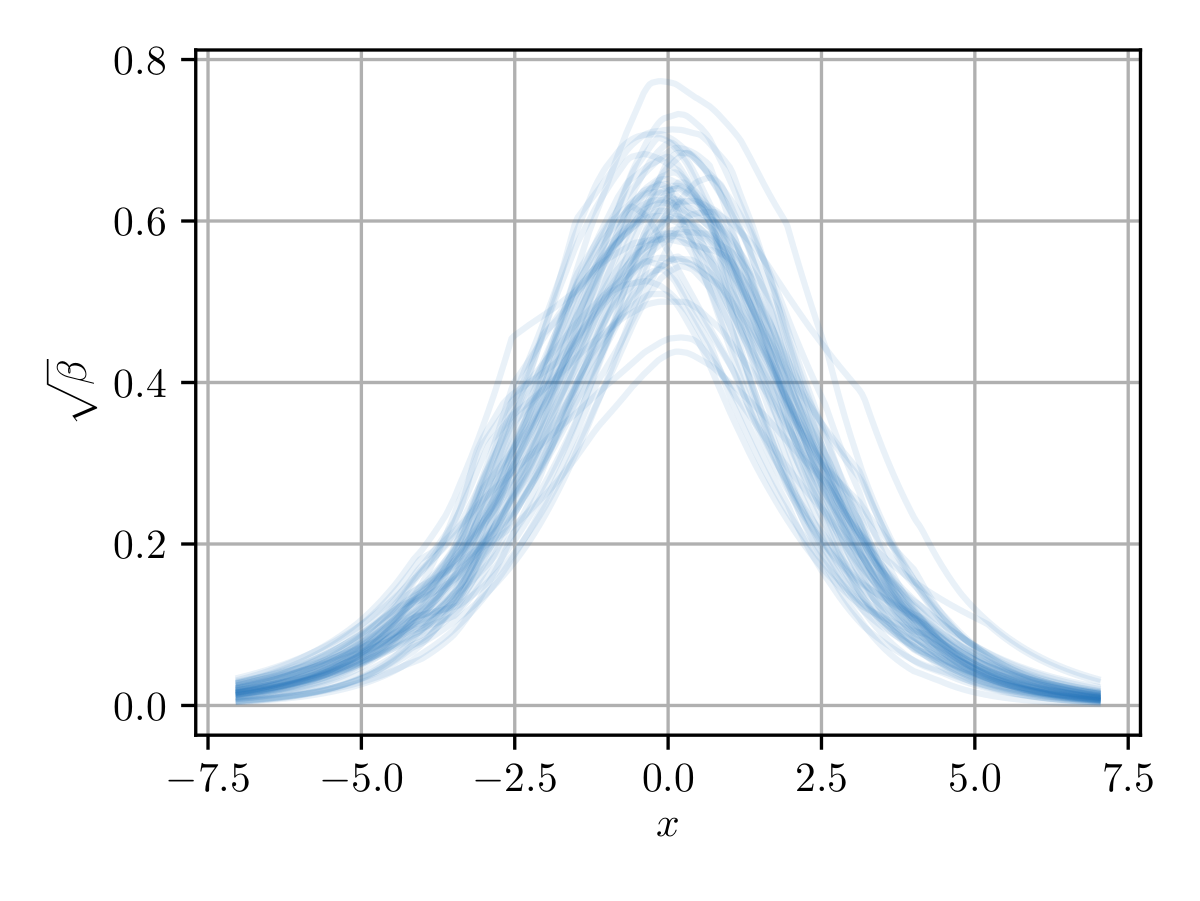}
        \caption{Distribution of $\beta$ in last epoch.}
        \label{fig:beta}
    \end{subfigure}
    \caption{Evolution of the parameters $\alpha_i$ and $\beta_i$. The distributions on the left are the result of averaging $\alpha_i$ and $\beta_i$ for 50 individually trained models after each epoch.}
\end{figure*}

\begin{figure*}[ht]
    \centering
    \begin{lstlisting}[numbers=none,caption={Patch for changing $u_\text{ep}$ to $u_\text{ep}'$ in https://github.com/aamini/evidential-deep-learning, Git commit hash d1d8e39},label={lst:patch_ep}]
diff --git a/neurips2020/gen_depth_results.py b/neurips2020/gen_depth_results.py
index a413288..a2c8ecc 100644
--- a/neurips2020/gen_depth_results.py
+++ b/neurips2020/gen_depth_results.py
@@ -435,3 +435,3 @@ def predict(method, model, x, n_samples=10):
         mu, v, alpha, beta = tf.split(outputs, 4, axis=-1)
-        sigma = tf.sqrt(beta/(v*(alpha-1)))
+        sigma = 1 / tf.sqrt(v)
         return mu, sigma
	\end{lstlisting}
	
    \begin{lstlisting}[numbers=none,caption={Patch for changing $u_\text{ep}$ to $u_\text{al}'$ in https://github.com/aamini/evidential-deep-learning, Git commit hash d1d8e39},label={lst:patch_al}]
diff --git a/neurips2020/gen_depth_results.py b/neurips2020/gen_depth_results.py
index a413288..5349195 100644
--- a/neurips2020/gen_depth_results.py
+++ b/neurips2020/gen_depth_results.py
@@ -435,3 +435,3 @@ def predict(method, model, x, n_samples=10):
         mu, v, alpha, beta = tf.split(outputs, 4, axis=-1)
-        sigma = tf.sqrt(beta/(v*(alpha-1)))
+        sigma = tf.sqrt(beta*(1+v)/v/alpha)
         return mu, sigma
    \end{lstlisting}
\end{figure*}

\end{document}